\documentclass{article}

\usepackage{arxiv}
\usepackage[utf8]{inputenc}
\usepackage[T1]{fontenc}
\usepackage{hyperref}
\usepackage{url}
\usepackage{booktabs}
\usepackage{amsmath, amsfonts}
\usepackage{nicefrac}
\usepackage{microtype}
\usepackage{graphicx}
\usepackage{lipsum}

\usepackage{cleveref}
\usepackage{authblk}
\usepackage{xcolor}
\usepackage{tabularx}
\usepackage{enumitem}
\usepackage{multirow}
\usepackage{array}
\usepackage{natbib}

\graphicspath{{./images/}}

\newcommand{\edit}[1]{\textcolor{black}{#1}}

\usepackage{setspace}
\usepackage[switch]{lineno}

\begin{document}

\begin{center}
{\Large \textbf{INQUIRE-Search: \\Interactive Discovery in Large-Scale Biodiversity Databases}}\\[1.0cm]
\end{center}


\noindent
\textbf{Edward Vendrow$^{1,*}$,
Julia Chae$^{1,*, \dagger}$,
Rupa Kurinchi-Vendhan$^{1,*, \dagger}$,
Isaac Eckert$^{2}$,
Jazlynn Hall$^{3}$,
Marta Jarzyna$^{4}$,
Reymond Miyajima$^{4}$,
Ruth Oliver$^{5}$,
Laura Pollock$^{2}$,
Lauren Shrack$^{1}$,
Scott Yanco$^{6}$,
Oisin Mac Aodha$^{7}$,
Sara Beery$^{1,\dagger}$}
\\

\noindent
$^{1}$Massachusetts Institute of Technology, Cambridge, MA, USA \\
$^{2}$McGill University, Montréal, QC, Canada \\
$^{3}$Cary Institute of Ecosystem Studies, Millbrook, NY, USA \\
$^{4}$The Ohio State University, Columbus, OH, USA \\
$^{5}$University of California Santa Barbara, Santa Barbara, CA, USA \\
$^{6}$Smithsonian's National Zoo \& Conservation Biology Institute, Washington, DC, USA \\
$^{7}$University of Edinburgh, Edinburgh, United Kingdom \\
$^{*}$Equal contribution \\
$^{\dagger}$Corresponding authors: \texttt{beery@mit.edu}, \texttt{chaenayo@mit.edu}, \texttt{rupak272@mit.edu} \\

\noindent
\section*{Abstract}
Many ecological questions center on complex phenomena, such as species
interactions, behaviors, phenology, and responses to disturbance, that are inherently
difficult to observe and sparsely documented. Community science platforms such as
iNaturalist contain hundreds of millions of biodiversity images, which often contain
evidence of these complex phenomena. However, current workflows that seek to
discover and analyze this evidence often rely on manual inspection, leaving this
information largely inaccessible at scale. We introduce INQUIRE-Search, an open-
source system that uses natural language to enable scientists to rapidly search within
an ecological image database like iNaturalist for specific phenomena, verify and export
relevant observations, and use these outputs for downstream scientific analysis.
Compared to existing methods, INQUIRE-Search concentrates relevant observations
3–8 times more efficiently under comparable manual inspection budgets across five
ecological case studies. This opening up new possibilities for scientific question
answering. Through five case studies, we demonstrate how INQUIRE-Search can be
used for ecological inference, from analyzing seasonal variation in behavior across
species to forest regrowth after wildfires. These examples illustrate a new paradigm for
interactive, efficient, and scalable scientific discovery that can begin to unlock
previously inaccessible scientific value in large-scale biodiversity datasets. Finally, we highlight how AI-enabled discovery tools for science require reframing aspects of the
scientific process, including experiment design, data collection, survey effort, and
uncertainty analysis. The code and data used in this study are available \href{https://github.com/Beery-Lab/INQUIRE-Search/tree/main}{here}.

\vspace{-2mm}
\newpage

\section{Introduction}
\vspace{-5mm}
Observations that capture \textit{ecological context beyond species occurrence}, such as species interactions and spatiotemporal changes in traits or habitat, are the fundamental underpinning of most ecological research and analyses. Significant resources are invested each year into capturing these observations via, e.g., transects or deployed sensors. Digital imagery and automated camera systems are an increasingly common mechanism for ecological observation \citep{oliver2023camera, kays2020born}. However, each image also records a broader ecological scene, incidentally capturing information about traits, habitat, seasonality, behavior, and interactions. For example, an image captured to document the occurrence of a bird species may also show the individual feeding, allowing seasonal or inter-species patterns in diet to be inferred; images collected to record species presence in recently disturbed landscapes may incidentally document early stages of forest regeneration; and repeated images of flowering or fruiting plants collected as occurrence records can provide fine-scale phenological information. Because these observations depend on organisms, conditions, and observers coinciding in time and space, they are difficult to collect systematically through targeted field studies.

Community-science platforms aggregate opportunistic observations at unprecedented spatial and temporal scales. Platforms such as iNaturalist, Pl@ntNet, and the Macaulay Library now host hundreds of millions of image-based observations worldwide \citep{iNaturalist2024, callaghan2021three, MacaulayLibrary}. 
{Together with broader open-access biodiversity infrastructures such as GBIF and Darwin Core \citep{gbif2019, wieczorek2012darwin}, these databases have transformed ecological research, but most applications still only rely on species occurrence information to support analyses of distributions, range limits, and broad-scale patterns of relative abundance \citep{eichholtzer2025integrating, song2025mapping, bosso2024integrating, moore2024leveraging, wijewardhana2022modelling, deshwal2021using}} \citep{kumar2019using, humphreys2019seasonal}.
Nevertheless, these unprecedented scales of data correspond to a higher likelihood of \textit{also} capturing diverse ecological phenomena from diet to phenology, which is often referred to as ``secondary data'' in the context of these repositories \citep{pernat2024overcoming, marques2024retrieving, davison2025automated}. \edit{As ecological image repositories such as iNaturalist continue to expand, an ecological informatics problem arises: how can we transform large, weakly indexed image repositories into curated datasets for targeted scientific questions?}

This inaccessibility stems from how observations are indexed and searched: existing filtering approaches rely heavily on basic metadata (e.g. taxonomic identity, location, and time), which describe the occurrence record but not its ecological context. Richer ecological descriptors (e.g. behavior, interactions, and habitat) are optional, inconsistently applied, or absent altogether; fewer than 25\%  of the over 300 million iNaturalist records contain any additional annotations \citep{iNaturalist2024}. For example, a recent study identifying predation events involving rove beetles required manually reviewing approximately 48,000 observations to locate just 159 relevant records \citep{hu2025global}.

Deep learning methods are well-posed to speed up large-scale image analysis. In ecology, supervised detection and classification models are already widely used to identify species in camera trap and community-science imagery \citep{norouzzadeh2018automatically, willi2019identifying} {\citep{de2025conformal, eichholtzer2025integrating}}.  
Training such supervised models to identify secondary ecological information---such as interactions, behaviors, or habitat context---requires concept-specific labeled datasets, specialized expertise, and substantial data and compute infrastructure, limiting their use for open-ended secondary ecological information at scale.

{Recent work has increasingly developed retrieval and mining tools for heterogeneous data, including unstructured text \citep{castro2024large}, and genomic sequence data \citep{rabelo2025datafishing}.} In the context of image data, vision-language models (VLMs) provide a promising path to reducing these access barriers by enabling flexible, efficient search over visual data with open-ended text, for example, letting an ecologist search all of iNaturalist for ``American Robin eating a worm.'' Trained on large-scale image-text datasets, VLMs learn to map images and natural language into a shared high-dimensional representation space, where semantically related concepts lie close together (Figure 1) \citep{radford2021learning}. This capability enables similarity-based retrieval of images depicting various contexts without requiring predefined labels. Recent work demonstrates the potential of this approach for ecological discovery: WildCLIP \citep{gabeff2024wildclip} fine-tuned a VLM on camera trap imagery and ecology-specific language, enabling the retrieval of previously unseen behaviors and habitat attributes from large wildlife datasets, while other related work has evaluated general-purpose vision-language models on expert-defined, open-ended ecological queries across millions of community-science images \citep{vendrow2024inquire}. \edit{However, these studies primarily establish retrieval capability. INQUIRE-Search builds on this work by showing how open-ended vision--language search can function within an expert-guided ecological workflow, from candidate discovery to curated datasets for downstream analysis.}

\begin{figure}[h]
    \centering
    \includegraphics[width=\textwidth]{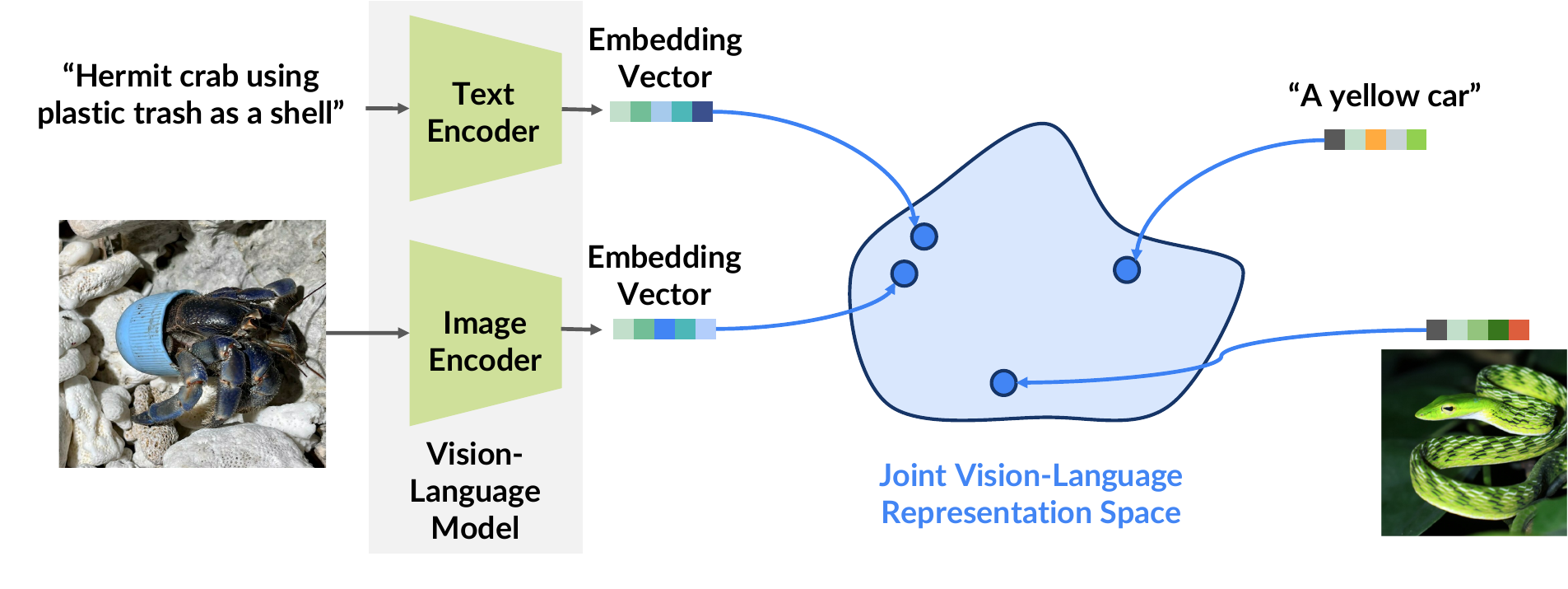}
    \caption{VLMs are trained to learn a joint embedding space between vision and text modalities, where similar semantic concepts are stored closer together. At search-time, the input text query is embedded by the text encoder of the VLM. Then the query embedding vector is compared against the pre-computed iNaturalist image embeddings to retrieve the most relevant images.}
    \label{fig:figure_1}
\end{figure}

\edit{Connecting large-scale image databases to ecological inference requires turning open-ended retrieval results into scientifically usable evidence through expert verification, metadata-rich export, and downstream evaluation.} In this work, we introduce INQUIRE-Search, an open-source, expert-driven system that enables efficient search over large ecological image databases using natural-language queries leveraging VLMs. INQUIRE-Search is designed as a discovery-oriented informatics system, treating ecological image analysis as an \textit{interactive information-retrieval task} rather than a closed-set predictive modeling problem. \edit{Our contributions are threefold:}
\begin{itemize}
     \item \edit{We introduce and evaluate INQUIRE-Search, an open-source ecological informatics framework that combines open-ended vision–language retrieval, expert verification, and iterative query refinement for downstream ecological analysis.}

    \item \edit{We show that \textbf{large community-science image repositories can be queried as sources of visual ecological evidence} for complex, context-dependent phenomena that are difficult to surface using metadata-based or closed-set approaches.}

     \item \edit{We \textbf{demonstrate the ecological utility of INQUIRE-Search} through diverse case studies, while characterizing practical limitations and future directions for retrieval-curated ecological evidence.}
\end{itemize}

\section{Methods and System Design}

\subsection{INQUIRE-Search system architecture}

INQUIRE-Search is an open-source system for interactive, expert-driven retrieval from large ecological image datasets. It integrates a state-of-the-art VLM, a high-performance vector index, and a lightweight, browser-based interface to support efficient natural-language search, verification, and data export.

\edit{To select the VLM backbone, we used INQUIRE-Bench, a subset of iNaturalist exhaustively annotated for ecological concepts and designed for ecological concept retrieval evaluation. Based on prior comparisons of state-of-the-art VLMs, we selected SigLIP-So400m-384-14 \citep{zhai2023sigmoid}, which achieved the strongest retrieval performance on this benchmark. INQUIRE-Search uses this vision-language model to map both images and natural-language queries into a shared semantic space and ranks images by similarity to the query.} All iNaturalist \citep{iNaturalist2024} images were preprocessed to a uniform format and embedded using the SigLIP image encoder. These embeddings were stored in a FAISS (Facebook AI Similarity Search) \citep{douze2025faiss} index, enabling sub-second similarity search across approximately 300 million images.

At search time, a user-provided text query is embedded and used to rank images in the FAISS index by cosine similarity. The browser-based interface supports metadata filtering (taxonomy, geography, and date), rapid inspection and expert verification of retrieved images, and export of curated observations with complete iNaturalist metadata. Verified image sets can be exported as CSV files, including observation IDs, coordinates, timestamps, taxonomic information, and file URLs, for downstream analysis. Full implementation details are provided in the \hyperref[sec:appendixa]{Appendix A}.

\subsection{INQUIRE-Search workflow}
\edit{We define ecological concept retrieval as ranking observations from a biodiversity image repository by how likely their images are to provide visually verifiable evidence of an ecological phenomenon described by a natural-language query. In this setting, relevance is determined by expert review using case-study-specific criteria.} We follow a standardized human-in-the-loop workflow using INQUIRE-Search (Figure 2).

\begin{enumerate}
    \item \textbf{Query and Prioritize:} We composed natural-language queries tailored to each ecological question and applied optional metadata filters (species, date, location). All prompts are listed in Table 1, with complete information on the filters, number of retrievals inspected, and number of samples marked for analysis provided in the \hyperref[sec:appendixb]{Appendix B}. 
    \item \textbf{Retrieve and Verify:} Each text query is embedded using the vision--language model and used to retrieve a ranked list of images from the vector index. {Images are inspected sequentially in descending similarity order. Across all retrieval tasks, images are labeled as informative only when the target ecological phenomenon is visually identifiable and unambiguous.}  Inspection budgets are defined ahead of time for each retrieval task (typically 200-500 images) to ensure comparable human effort across evaluations. {In some cases, inspection terminated early when extended sequences of non-relevant images indicated declining relevance in the ranked results.} 
    \item \textbf{Export and Analyze:} Informative images and their associated metadata are exported in tabular (CSV) format for downstream analysis. Task-specific analytical procedures are described in Section~\ref{sec:case_studies}.

\end{enumerate}

\begin{figure}[h]
    \centering
    \includegraphics[width=\textwidth]{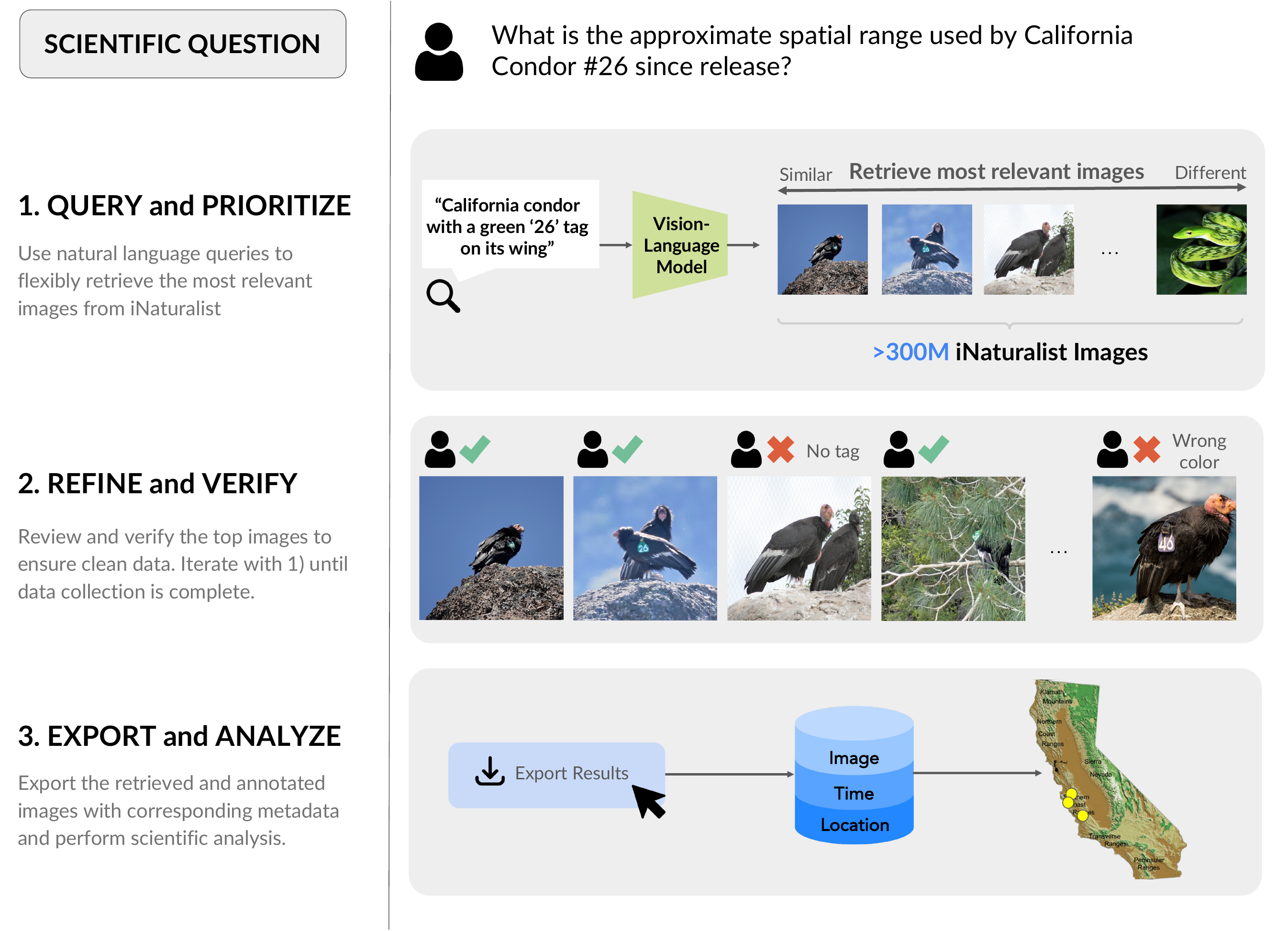}
    \caption{\textbf{INQUIRE-Search pipeline overview.} Users begin with a scientific question and iteratively (1) query and prioritize relevant iNaturalist images, (2) refine and verify results with expert inspection, and (3) export the curated data for downstream analysis. The process supports returning to earlier stages as needed until sufficient data are collected.}
    \label{fig:figure_2}
\end{figure}

\begin{table}[ht]
\centering
\small
\begin{tabularx}{\textwidth}{l X}
\toprule
\textbf{Case study} & \textbf{INQUIRE-Search Prompt} \\
\midrule

Seasonal variation in bird diets &
Prompts of the form ``\textbf{<species>} with \textbf{<diet type>} in its mouth'' using:
\begin{itemize}[noitemsep, topsep=0pt, leftmargin=*]
    \item \textbf{Species} (common names): Gray-cheeked Thrush, Ancient Murrelet, American Tree Sparrow, Red-bellied Woodpecker, American Robin
    \item \textbf{Diet types}: invertebrate, vertebrate, seed, fruit, nectar and pollen, carrion or animal decay, other plant matter
\end{itemize}
Total: 5 species × 7 diet types = 35 prompts. \\
\midrule

Post-fire forest regrowth &
``Young coniferous trees in burned forest'' \\
& ``Young deciduous trees in burned forest'' \\
\midrule

Wildlife mortality &
``Dead bird'' \\
\midrule

Plant Phenology &
``Milkweed germinating or emerging'' \\
& ``Milkweed flowering'' \\
& ``Milkweed producing seeds'' \\
& ``Milkweed dying or withering / in senescence'' \\
\midrule

Whale re-identification &
``White underside of humpback whale fluke'' \\
\bottomrule
\end{tabularx}
\vspace{5pt}
\caption{Prompts used for retrieval across the five INQUIRE-Search case studies.}
\label{tab:standard_prompts}
\end{table}

\subsection{Benchmarking and evaluation}
{For each retrieval task corresponding to a specific scientific question}, we evaluate INQUIRE-Search along two complementary directions: 1) retrieval efficiency and 2) ecological utility.

\textit{\textbf{Retrieval efficiency}} is evaluated using standardized, effort-oriented metrics appropriate for open-ended discovery \edit{over target phenomena that are long-tailed, inconsistently annotated, or not known a priori.} Under these conditions, traditional accuracy-based metrics such as precision, recall, or mean average precision are not only infeasible but potentially misleading. For each search, we record the number of retrieved images manually inspected ($N_{\text{insp}}$), the number of images verified as informative and retained for analysis ($N_{\text{ret}}$; final dataset size), and the screening yield ($Y = N_{\text{ret}} / N_{\text{insp}}$). However, screening yield is not interpreted as a measure of model performance or retrieval accuracy, since the prevalence of relevant observations varies across queries and study contexts, but is used comparatively to assess how efficiently INQUIRE-Search concentrates relevant observations under comparable human effort for inspection. \edit{For case studies with comparable ranked outputs, we also plot recovery curves showing the cumulative number of verified informative images recovered as a function of inspection depth. These curves evaluate whether relevant observations are concentrated near the top of the ranking and how efficiency changes with additional review effort.}

To enable such comparison, we evaluate INQUIRE-Search against a traditional iNaturalist filtering baseline representing best-practice metadata-based search. Baseline queries apply the same taxonomic, spatial, and temporal filters used in INQUIRE-Search to ensure comparable search scope. Beyond these shared constraints, baselines rely on structured metadata fields when relevant controlled attributes exist, and otherwise use keyword matching over unstructured text fields such as observation descriptions, tags, and annotations. Keyword queries are constructed to approximate the semantic intent of the corresponding INQUIRE-Search prompts as closely as possible. All retrieved results are screened using identical expert verification criteria. Where applicable, we report the yield ratio ($Y_{\text{ratio}} = Y_{\text{INQUIRE-Search}} / Y_{\text{baseline}}$), which captures the relative efficiency between INQUIRE-Search and the baseline method under fixed inspection budgets. All keyword queries are specified within the Results of each case study.

\textit{\textbf{Ecological utility}} is assessed by examining whether the datasets produced by INQUIRE-Search for a given retrieval task are sufficient to support meaningful downstream ecological analyses. These analyses are described in Section~\ref{sec:case_studies}.

\subsection{Data availability and reproducibility}
{INQUIRE-Search is released as an open-source system, and the full codebase, documentation, and instructions for reproducing all experiments are available at \url{https://github.com/Beery-Lab/INQUIRE-Search}. Case study datasets are derived from openly accessible iNaturalist observations and are exported with complete associated metadata (including observation identifiers, coordinates, timestamps, taxonomic information, and licenses) to support transparency and reproducibility.}


\section{Validation with Ecological Case Studies}
\label{sec:case_studies}

{To illustrate how this methodology could support ecological analyses, we present a set of preliminary, scientifically diverse case studies.} We first identified questions that require contextual information---such as behaviors, interactions, or life-history events. We focus on phenomena that are likely to be present {and visible} in community-science images but are rarely annotated, and therefore difficult to access at scale. \textbf{{Each case study is designed to evaluate the ability of INQUIRE-Search to retrieve relevant data, rather than to conduct a full ecological study.}} Using INQUIRE-Search, we recovered and analyzed relevant observations for each question, designing each experiment {to evaluate the usability, flexibility, and limitations of the tool rather than to produce definitive scientific conclusions.} Each case study explores a distinct question: (1) seasonal variation in bird diets, (2) post-fire forest regeneration, (3) spatio-temporal patterns of wildlife mortality, (4) plant phenology across seasonal cycles, and (5) individual Re-ID in humpback whales. These examples demonstrate the validity and diversity of carefully designed natural language-guided image search as a novel mechanism for \textit{data collection}, providing a scalable and flexible approach for extracting ecological insights from large, unstructured image datasets.

\subsection{CS1: Seasonal variation in bird diets}

\textbf{Overview.} {This case study evaluates whether INQUIRE-Search can recover feeding events across seasons from community-sourced photographs and reproduce known dietary patterns. We compare the trends in INQUIRE-Search retrieved dataset against published dietary records in SAviTraits \citep{murphy2023SAviTraits}.}

Diet is a fundamental response and effect trait that shapes vertebrate survival, fitness, trophic position, and ecological interactions, and it varies widely across species and environmental conditions \citep{burin2016omnivory, sibly2012energetics, belmaker2012global}. Although large dietary databases exist \citep{wilman2014eltontraits}, seasonal diet information remains sparse: in SAviTraits 1.0 \citep{murphy2023SAviTraits}, only about 10\% of more than 10,000 bird species are recorded as exhibiting seasonal dietary shifts. Community-science image repositories may fill these gaps by capturing feeding events as secondary data, including seasons and contexts that are difficult to sample systematically through targeted field studies. We try to surface these observations efficiently using INQUIRE-Search. In this case study, we focus on five species---Gray-cheeked Thrush (\textit{Catharus minimus}), Ancient Murrelet (\textit{Synthliboramphus antiquus}), American Tree Sparrow (\textit{Spizelloides arborea}), Red-bellied Woodpecker (\textit{Melanerpes carolinus}), and American Robin (\textit{Turdus migratorius})---which have high dietary certainty scores in SAviTraits \citep{murphy2023SAviTraits}, providing a robust reference for comparison.

\textbf{{INQUIRE-Search workflow for CS1.}} 
{\textbf{1) Query and Prioritize:}} For each target species, we construct natural language search queries specifying the bird species and diet type (e.g., ``\textit{Turdus migratorius} with invertebrate in its mouth''), using the dietary classification system used in SAviTraits. This classification system includes seven dietary categories: (1) invertebrate, (2) vertebrate, (3) seed, (4) fruit, (5) nectar and pollen, (6) carrion or animal decay, and (7) other plant matter. Each diet category was queried separately for summer (June-August) and winter (December-February). Taxonomic filters restricted results to the target species. {\textbf{2) Retrieve and Verify:}} The top 500 images for each species-diet type-season combination were screened by a reviewer to filter for images with a clearly identifiable food item visible in the bird's bill, with maximum of $N_{insp}=7000$ per species (some species had fewer than 500 observations for a specific season, resulting in smaller $N_{insp}$). {\textbf{3) Export and Analyze:}} For each combination, we counted the marked images and calculated the proportion of informative images belonging to each diet category to compare qualitatively with published seasonal diet compositions from SAviTraits. When comparing retrieval across species, we compare $N_{ret}$.

\textbf{{Results.}} Across the five study species, INQUIRE-Search revealed substantial variation in the availability of dietary information in community-science platforms (Table \ref{tab:cs1_birds}). Aggregated across diet categories and seasons, INQUIRE-Search retained the highest number of informative images ($N_{ret}= 669$) for American Robin (\textit{Turdus migratorius}), followed by Red-bellied Woodpecker (\textit{Melanerpes carolinus}) and American Tree Sparrow (\textit{Spizelloides arborea}) ($N_{ret}= 395$ and $188$, respectively), while for both Ancient Murrelet (\textit{Synthliboramphus antiquus}) and Gray-cheeked Thrush (\textit{Catharus minimus}) we found a negligible number of images containing dietary information ($N_{ret} <= 2$).

Notably, standard iNaturalist search tools recovered markedly fewer diet observations. Keyword-based queries (i.e. species filtering and using ``eating'' + diet term as a keyword search) returned $N_{ret}= 105$ feeding observations for American Robin, $27$ for Red-bellied Woodpecker, and $12$ for American Tree Sparrow, with zero results for the remaining species. In many cases, baseline queries returned fewer total images than the inspection budget, indicating that diet-related feeding events are sparsely indexed in structured metadata and text fields. In practice, these are insufficient for seasonal comparison.

\begin{figure}[h]
    \centering
    \includegraphics[width=\textwidth]{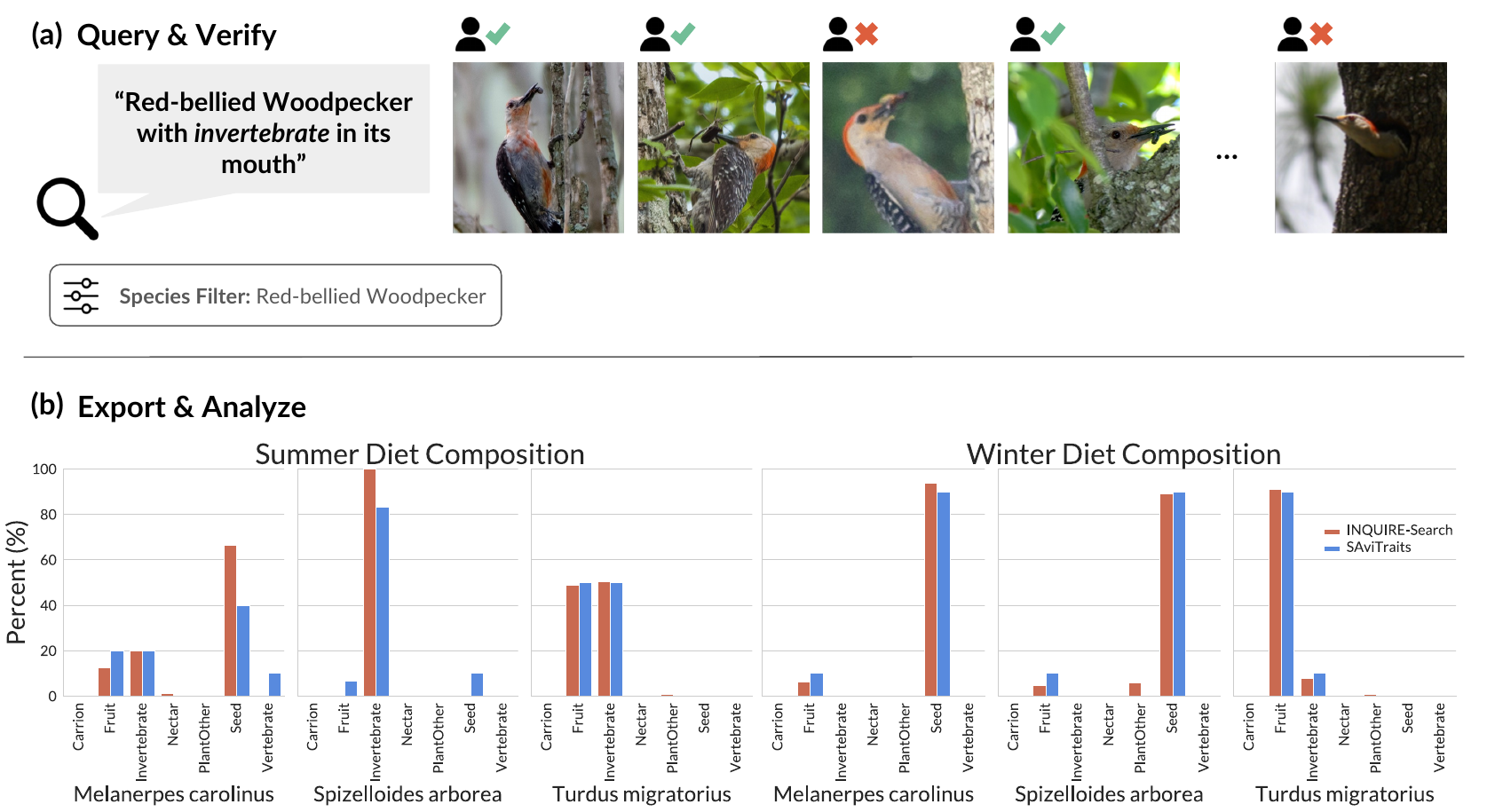}
    \caption{\textbf{(Top) Query \& Verification.} Filtering top INQUIRE-Search outputs for ``Red-bellied Woodpecker'' and ``Invertebrate.'' \textbf{(Bottom) Diet comparisons.} INQUIRE-Search yields dietary patterns that align closely with those documented in the SAviTraits reference dataset.}
    \label{fig:figure_3}
\end{figure}

Where sufficient data were available, INQUIRE-Search results closely match known seasonal diet compositions reported in SAviTraits, as seen in Figure 3, with agreement in both dominant food types and their relative compositions. For example, American Robin summer diets were composed of roughly equal proportions of fruits and invertebrates, while Red-bellied Woodpecker showed a diet dominated by seeds (50\% in INQUIRE-Search vs. 40\% in SAviTraits). In winter, all three analyzable species exhibited diets strongly dominated by a single food category, reflecting increased specialization during colder months. 

These results show that INQUIRE-Search can recover large numbers of verifiable feeding events and reproduce known seasonal diet trends, provided that the focal species is frequently photographed within community-science repositories. Two of the species, however, lacked sufficient image data for meaningful analysis despite being well-studied in the scientific literature and represented in databases like SAviTraits. This limitation is likely even more pronounced for understudied species---precisely those for which new dietary data are most needed---because they are also underrepresented in iNaturalist.

\subsection{CS2: Post-fire forest regrowth}

\textbf{Overview.} {This case study evaluates whether INQUIRE-Search can identify young coniferous and deciduous individuals within the 2012 High Park Fire perimeter from community-sourced photographs and examine how their occurrence varies across burn severity classes (Figure 4).}

\begin{figure}[h]
    \centering
    \includegraphics[width=\textwidth]{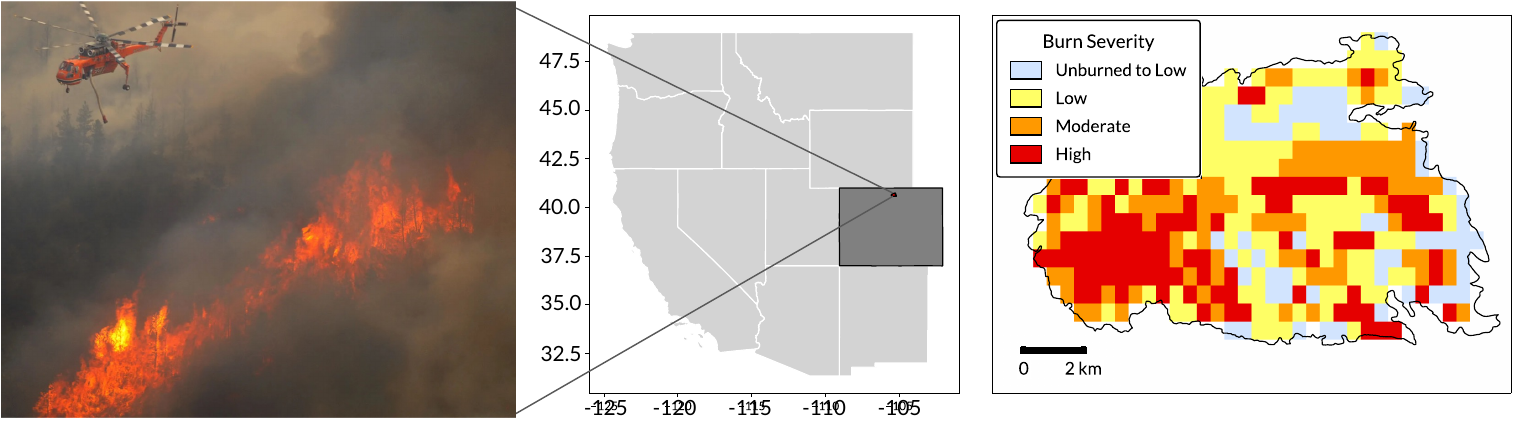}
    \caption{\textbf{(Left) Forest regrowth following the 2012 High Park Fire in Colorado.} With wildfires becoming more frequent and severe, fine-scale characterization of vegetation recovery is critical for understanding forest resilience. \textbf{(Right) MTBS burn severity categories.} We aggregate the burn categories from 30m Landsat resolution to a spatial resolution matching iNaturalist metadata (0.01$^{\circ}$, $\sim$1100m).}
    \label{fig:figure_4}
\end{figure}


Recent warming and drying trends associated with climate change are driving wildfires that are more frequent, severe, and intense \citep{abatzoglou2016impact, juang2022rapid}, exceeding the adaptive capacity of many forest systems \citep{harvey2016burn, davis2023reduced}. Understanding post-fire recovery is essential to assessing long-term forest resilience, successional trajectories, and long-term ecosystem change \citep{whitlock2008long, marlon2012long, keeley2011fire}. However, existing approaches limit the study of fine-scale regrowth: satellite data often lack the spatial or spectral resolution needed to detect young trees or differentiate species \citep{xu2021spatial, kiel2022trees}, while field surveys are constrained by cost and limited spatial coverage. In contrast, community-sourced photographs can reveal early regeneration that is invisible to both satellites and sparse field campaigns.

\textbf{{INQUIRE-Search workflow for CS2.}} \textbf{{1) Query and Prioritize:}} We targeted two functional groups (young coniferous trees and young deciduous trees), using text queries such as ``young coniferous trees in burned forest'' and ``young deciduous trees in burned forest.'' Spatial filters restricted observations to the High Park Fire perimeter:  latitude (40.57 - 40.75$^{\circ}$ N) and longitude (105.18 - 105.54$^{\circ}$ W) (Figure 5). Temporal filters restricted images to post-fire dates. After retrieval, we applied a finer spatial filter using reported image coordinates, retaining only images within the High Park MTBS fire perimeter via the \texttt{sf} package in R$^3$ \citep{pebesma2018simple}. {\textbf{2) Retrieve and Verify:}} From the top 200 images retrieved for each functional group, we marked images that clearly contained at least one individual tree of the target forest type. We allowed individual images to be marked as informative in multiple searches if the image contained individuals representing both forest types (e.g. a coniferous and deciduous tree seedling). We also included trees and shrubs within the same target forest type as it was difficult to distinguish young trees from shrubs in images. {\textbf{3) Export and Analyze:}} Coordinates of informative images were combined with MTBS burn severity categories by aggregating the 30 m burn data to the 0.01$^{\circ}$  precision typical of iNaturalist coordinates and assigning each location the modal severity class using the \texttt{terra} package in R$^4$ \citep{hijmans2022package}. We then counted marked images by forest type and burn severity to evaluate how regeneration outcomes varied across burn categories.

\textbf{{Results.}} INQUIRE-Search successfully identified photographs depicting early recovery within the High Park Fire boundary (Table \ref{tab:cs2_forest}). After fine-grained spatial filtering, we reviewed $N_{insp}=112$ deciduous and $N_{insp}=100$ coniferous images, verifying $N_{ret}=78$ deciduous ($Y=0.70$) and $N_{ret}=45$ coniferous ($Y=0.45$) regeneration events.

Without INQUIRE-Search, baseline retrieval relied on geographic and temporal filtering, followed by inspection of the first $N_{insp}=200$ images within the fire boundary. This yielded only $N_{ret}=6; Y=0.03$ coniferous and $N_{ret}=19; Y=0.095$ deciduous shrubs or seedlings. \edit{The recovery curves in Figure 6 show that INQUIRE-Search concentrates relevant post-fire recovery images early in inspection, matching the baseline’s full 200-image yield after only 9 inspections for coniferous recovery and 46 for deciduous recovery. Its continued upward trajectory further suggests that additional inspection effort would likely keep expanding the usable recovery dataset.} Under comparable manual inspection budgets, INQUIRE-Search produced substantially denser datasets, enabling subsequent analysis of regeneration patterns across burn-severity gradients. 

\begin{figure}[h]
    \centering
    \includegraphics[width=\textwidth]{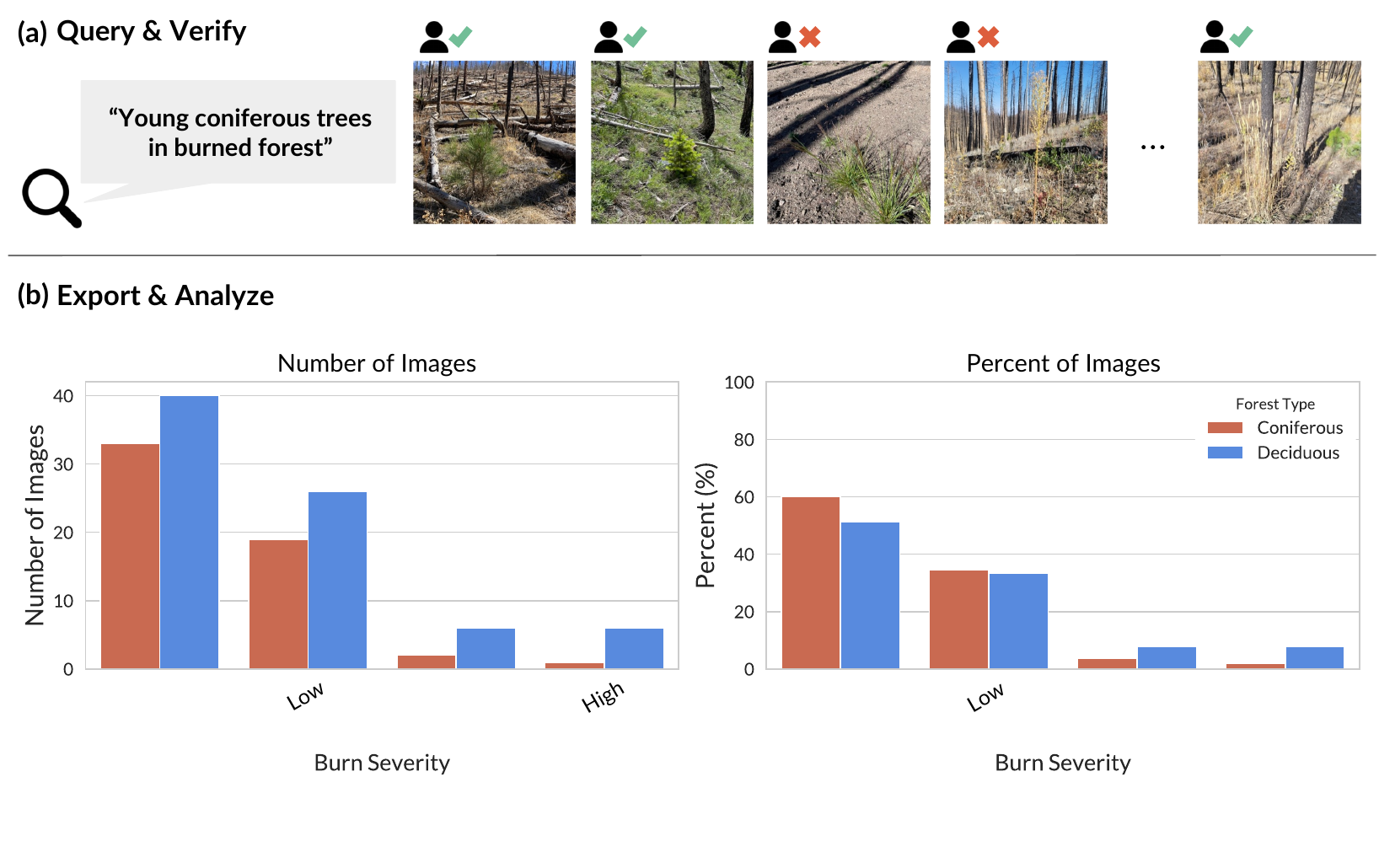}
    \caption{\textbf{(Top) Query \& Verification.} Filtering top INQUIRE-Search returns for ``young coniferous trees in burned forest.'' \textbf{(Bottom) Tree observations across burn severity.} Distribution of observed young coniferous and deciduous trees in community-science collected images show strong relationships with burn severity regions.}
    \label{fig:figure_5}
    \vspace{-4mm}
\end{figure}

\begin{figure}[h]
    \centering
    \includegraphics[width=\textwidth]{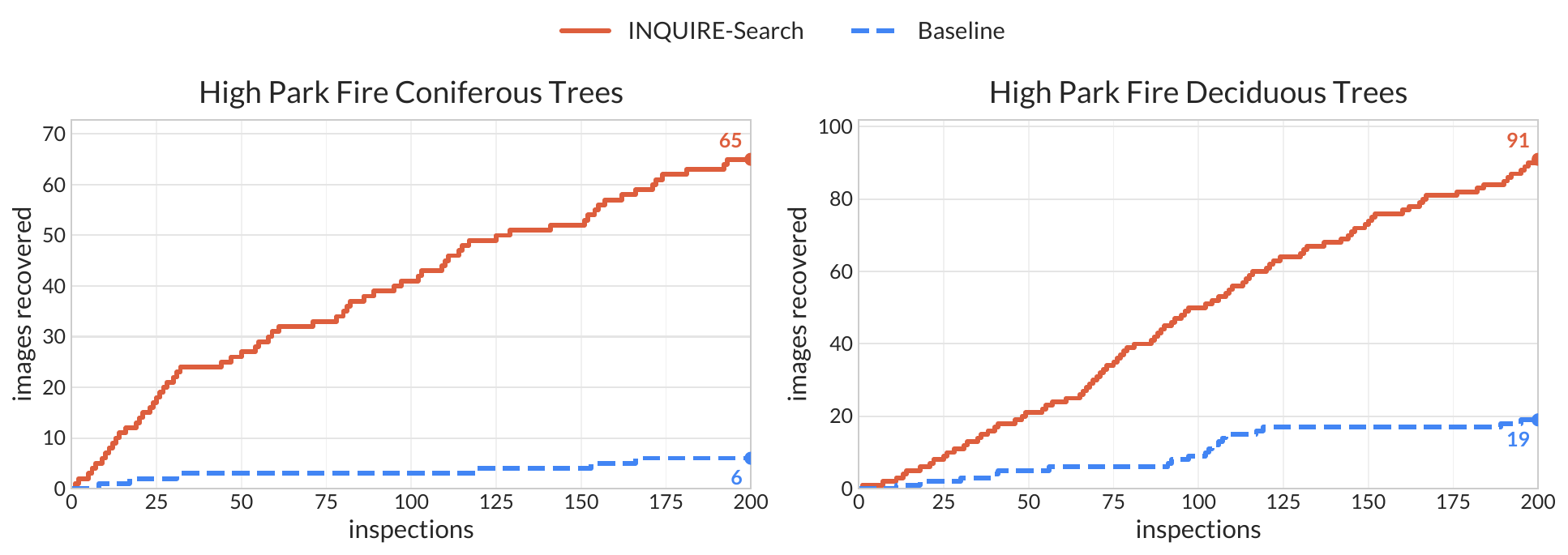}
    \caption{\textbf{Retrieval Curves for Fire Recovery.} Comparing recovered images vs inspections show that INQUIRE-Search is both significantly more efficient and effective at prioritizing relevant images for data collection.}
    \label{fig:fig_6}
    \vspace{-4mm}
\end{figure}

Using the curated INQUIRE-Search dataset, we found strong relationships between burn severity and post-fire regeneration. However, the strength of this pattern differed between functional groups. Coniferous regeneration showed a stronger negative association with increasing burn severity, while deciduous regeneration was somewhat more evenly distributed (Figure 5b, right). These patterns are consistent with previous research documenting that increased fire severity decreases conifer regeneration likelihood following the High Park fire and across regional forests \citep{davis2023reduced, wright2017post}. 

Several limitations remain, including coarse coordinate precision (image coordinates are rounded to 0.01$^{\circ}$ to match the limiting spatial precision of many iNaturalist records), challenges distinguishing seedlings and shrubs in photographs, and biases toward accessible locations, which may underrepresent severely burned patches (Figure 4). 

\subsection{CS3: Wildlife mortality}

\textbf{Overview.} {This case study explores the use of INQUIRE-Search to collect mortality instances in urban (Boston, MA) and rural (Pioneer Valley, MA) settings across time to investigate seasonal mortality dynamics and to compare relative mortality across different anthropogenic contexts (Figure 7).}

\begin{figure}[h]
    \centering
    \includegraphics[width=\textwidth]{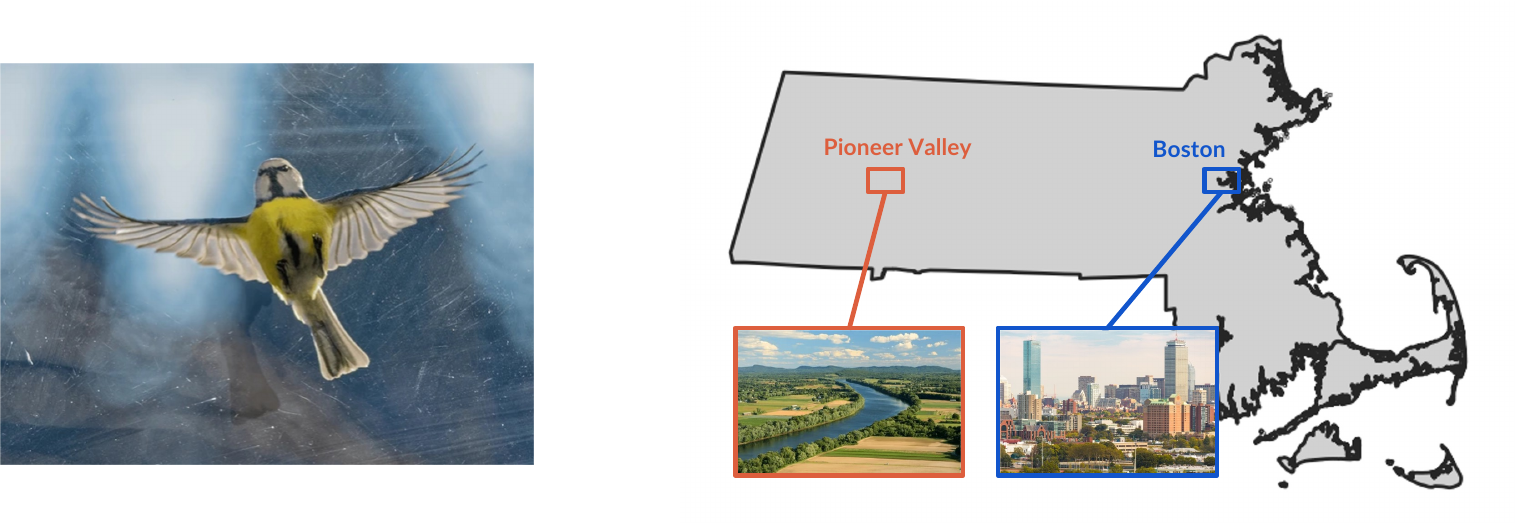}
    \caption{\textbf{(Left) Wildlife mortality trends vary by season and location.} Globally, collisions with human-made structures such as windows are a major cause of bird deaths. \textbf{(Right) Comparing avian mortality in rural vs urban regions.} Using INQUIRE-Search, we compared image-based evidence of avian mortality between urban (Boston, MA) and rural (Pioneer Valley, MA) sites to examine how seasonal risks differ across anthropogenic contexts.}
    \label{fig:figure_7}
    \vspace{-4mm}
\end{figure}

Understanding avian mortality is central to conservation, particularly given the loss of nearly three billion birds in North America since 1970 \citep{rosenberg2019decline}. Mortality risk varies seasonally \citep{loss2015direct, marra2015call} and is strongly influenced by collisions with human-made structures, including windows and vehicles \citep{loss2014estimation}, which peak during migration \citep{riding2021multi, scott2023causes} and vary across landscapes with bird abundance and anthropogenic context \citep{hager2017continent}. Yet documenting spatiotemporal mortality patterns is difficult because events are rare, and alternatives such as marking or tracking individuals are resource-intensive and limited in scale \citep{yanco2025tracking}. Because community-science observations are often collected near human activity, they frequently capture wildlife mortality events that would otherwise go unrecorded.

\textbf{{INQUIRE-Search workflow for CS3.}} \textbf{{1) Query and Prioritize:}} We queried for ``dead bird'' within two equal-sized bounding boxes using location filters: an urban site (Boston) (42.31 to 42.38$^\circ$ N, -71.14 to -71.01$^\circ$ W) and a rural site (Pioneer Valley) (42.31 to 42.38$^\circ$ N, -72.64 to -72.51$^\circ$ W). \textbf{{2) Retrieve and Verify:}} To identify true avian mortality events, we excluded images of live birds, single anatomical features, non-avian taxa, and duplicate observations of the same event (defined as records sharing species, month, latitude, and longitude). Images were reviewed sequentially in ranked order, and verification terminated after 200 consecutive non-relevant retrievals. Using this criterion, we verified $N_{insp}=359$ images in Pioneer Valley. In Boston, where upload density was substantially higher, we capped verification at $N_{insp}=1000$ retrievals due to a limited annotation budget, although additional relevant observations may have been present beyond this threshold. \textbf{{3) Export and Analyze:}} To control for differing bird abundance, observation effort, and mortality among sites, we calculate the \textit{mortality index} defined as: 
\[
\text{MortalityIndex}_{m,s} =
\log_{2}\!\left(
\frac{R_{m,s}}{\overline{R}_{s}}
\right),
\]
where the monthly mortality rate is
\[
R_{m,s} = \frac{D_{m,s}}{O_{m,s}},
\]
and the mean monthly mortality rate for site $s$ is
\[
\overline{R}_{s} = \frac{1}{12} \sum_{m=1}^{12} R_{m,s}.
\]
Here, $D_{m,s}$ is the mortality count and $O_{m,s}$ is the observation count for month $m$ at site $s$. Values above zero indicate an increase in mortality relative to the site's annual mean, while values below zero indicate a decrease. The total observation count was extracted using the \texttt{rinat} package in R \citep{barve2022rinat}.

\textbf{{Results.}} Table 
\ref{tab:cs3_mortality} show that INQUIRE-Search identified hundreds of verified avian mortality events in both rural and urban settings. In Boston, inspection of $N_{insp}=1000$ yielded $N_{ret}=545$ and in Pioneer Valley, $N_{ret}=79$ events were verified from $N_{insp}=360$ images. 

\begin{figure}[h]
    \centering
    \includegraphics[width=\textwidth]{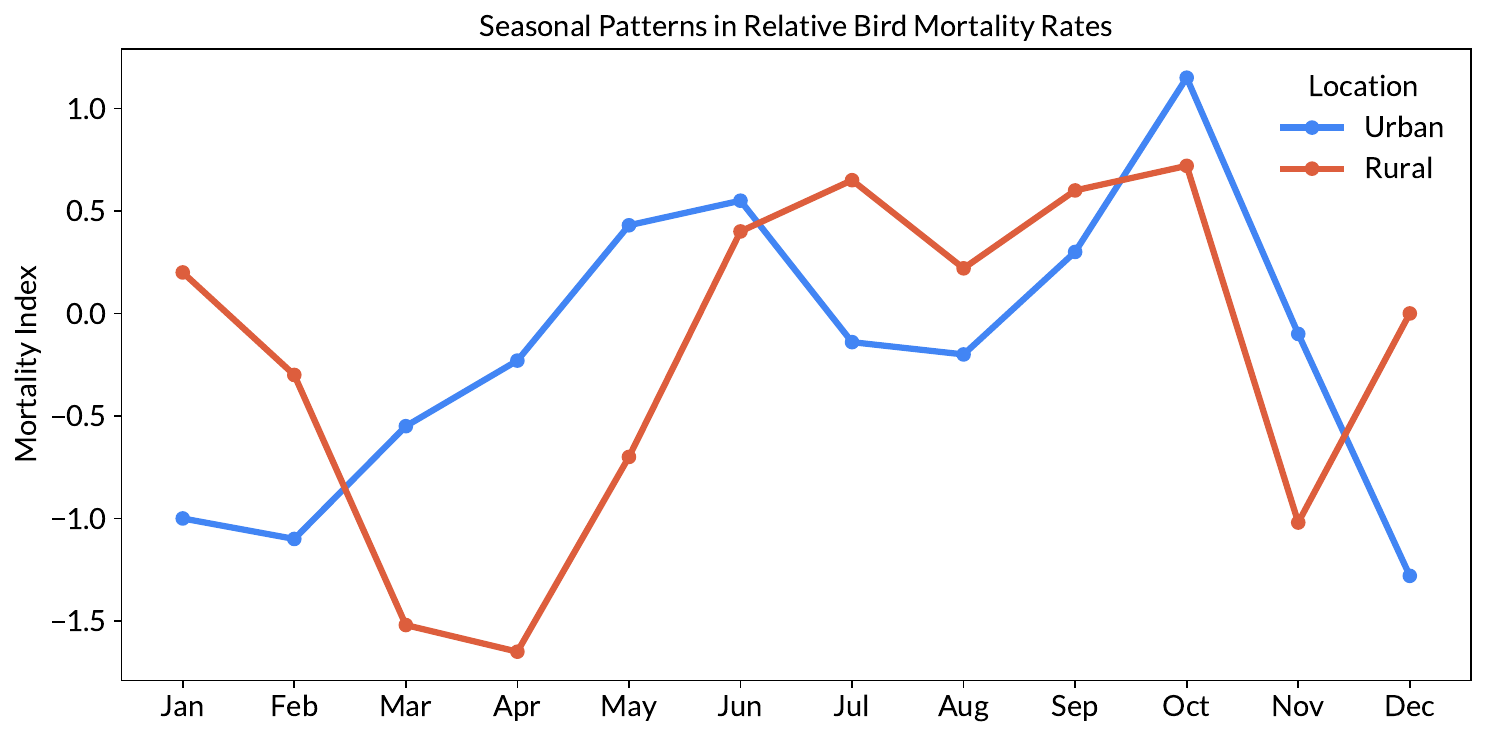}
    \caption{\textbf{Seasonal patterns in relative bird mortality for an urban (Boston, blue) and rural (Pioneer Valley, orange) region of Massachusetts}. Mortality is shown as a $\log_2$ index scaled to each site’s annual mean (0), with units representing fold-change. Positive values indicate above-average mortality, negative values below.}
    \label{fig:figure_8}
\end{figure}

In comparison, existing iNaturalist filters recovered significantly fewer mortality events. iNaturalist includes structured annotations for ``dead/alive'' attributes that observers can optionally provide. However, applying this filter yielded only $N_{ret}=295$ dead-bird observations in Boston and $N_{ret}=42$ in Pioneer Valley. Using keyword queries performed even worse, returning just $N_{ret}=46$ and $N_{ret}=8$ candidate images in Boston and Pioneer Valley respectively. In many cases, baseline searches returned fewer total images than the nominal inspection budget, indicating that mortality events are sparsely indexed in structured metadata and text fields and are difficult to recover using conventional metadata- and keyword-based search alone.

Using the verified datasets, we examined seasonal mortality dynamics across the two landscapes (Figure 8). Both sites exhibited high mortality during fall migration (September-October), consistent with increased collision risk in the eastern United States \citep{horton2019bright}. Interestingly, Boston exhibited higher mortality rates during spring migration, potentially reflecting differences in migratory passage rates or site-specific risk factors \citep{scott2023causes}. In contrast, Pioneer Valley lacked a pronounced winter decline, which may reflect differences in community composition or potentially increased winter detectability due to reduced vegetation, which would disproportionally affect rural data collection. 

Overall, these findings show that INQUIRE-Search can detect distinct spatiotemporal patterns in avian mortality that are difficult to capture using metadata alone. The contrasting seasonal dynamics between urban and rural sites highlight the role of anthropogenic context in driving avian demographics. However, interpretation must account for detection biases: urban carcasses are more visible, while rural mortality may be underrepresented due to scavenging and remoteness. Additionally, our mortality index assumes that the ratio of dead bird observations to total bird observations provides a meaningful proxy for actual mortality rates, but this assumption may not hold if detection probabilities vary systematically across seasons, locations, or species.

\subsection{CS4: Resolving plant phenology}

\textbf{Overview.} {In this case study, we test whether INQUIRE-Search can recover four distinct phenophases---emergence, flowering, seeding, and senescence---for common milkweed (\textit{Asclepias syriaca}) in southern Québec from community-science images (Figure 9).}

\begin{figure}[h]
    \centering
    \includegraphics[width=\textwidth]{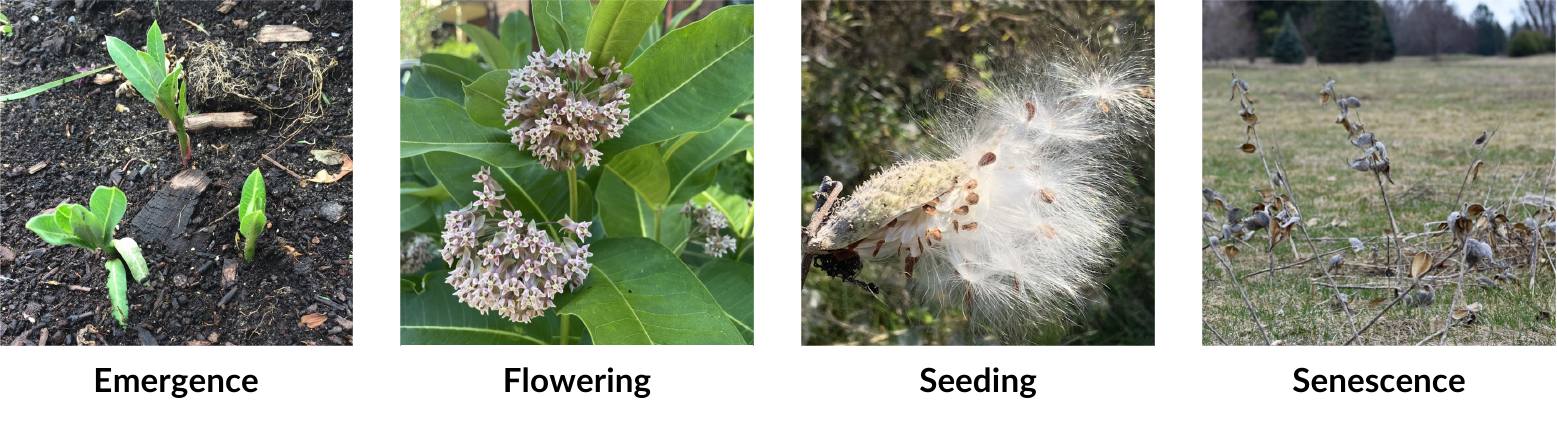}
    \caption{\textbf{Four phenological stages of common milkweed.} Example observations identified using INQUIRE-Search that correspond to the four phenological stages analyzed in this study: emergence, flowering, seeding, and senescence.}
    \label{fig:figure_9}
\end{figure}

Studying life-history timing is a primary way researchers detect biodiversity responses to global change \citep{parmesan2003globally}. While fine-scale phenological datasets from repeated field surveys provide local insight \citep{austin2024climate}, phenological change remains difficult to assess at broader spatial scales \citep{doi2017macroecological}. Although some studies use computer vision to extract phenological information \citep{williamson2025long}, most rely on coarse metrics, such as mean flowering date, because data on distinct phenophases are sparse. The growing volume of plant observations on community-science platforms like iNaturalist offers a path to overcoming these limitations, provided phenological stages can be identified from photographs.

\textbf{{INQUIRE-Search workflow for CS4.}} \textbf{{1) Query and Prioritize:}} We created stage-specific text queries targeting each phase:  (1) emergence (``Milkweed germinating or emerging''), (2) flowering (``Milkweed flowering''), (3) seeding (``Milkweed producing seeds or milkweed with seeds''), and (4) senescence (``Milkweed dying or withering or senescence''). We also used a species filter (``\textit{Asclepias syriaca}'') and geographic filters for latitude (45.03 - 46.54$^{\circ}$ N) and longitude (-74.68 - -71.66$^{\circ}$ W) to limit observations to southern Quebec. \textbf{{2) Retrieve and Verify:}} The top $N_{insp}=200$ retrieved images were manually verified and labeled as informative based on strict, visually identifiable morphological criteria corresponding to each phenophase. Images were classified as emergence if plants had fewer than four pairs of adult leaves; as flowering if open petals were visible; as seeding if seed pods were open with clearly visible seeds; and as senescence if green leaves were absent. \textbf{{3) Export and Analyze:}} Observation dates were converted to Day-of-Year (DOY). Mean DOY values were compared among stages using ANOVA, followed by Tukey’s HSD to identify pairwise differences.

\textbf{{Results.}}
INQUIRE-Search retrieved stage-specific phenological observations of common milkweed in southern Quebec, with verification success varying strongly by phenophase (Figure 10). Flowering and seeding yielded the largest datasets ($N_{ret}=169$, $Y=0.85$; $N_{ret}=161$, $Y=0.81$), while emergence and senescence produced smaller but still substantial datasets ($N_{ret}=45$, $Y=0.23$; $N_{ret}=52$, $Y=0.26$).  As illustrated by the verified examples in Figure 10, these differences likely reflect both the visibility of diagnostic morphological cues and observer preferences for photographing conspicuous life stages.

\begin{figure}[h]
    \centering
    \includegraphics[width=\textwidth]{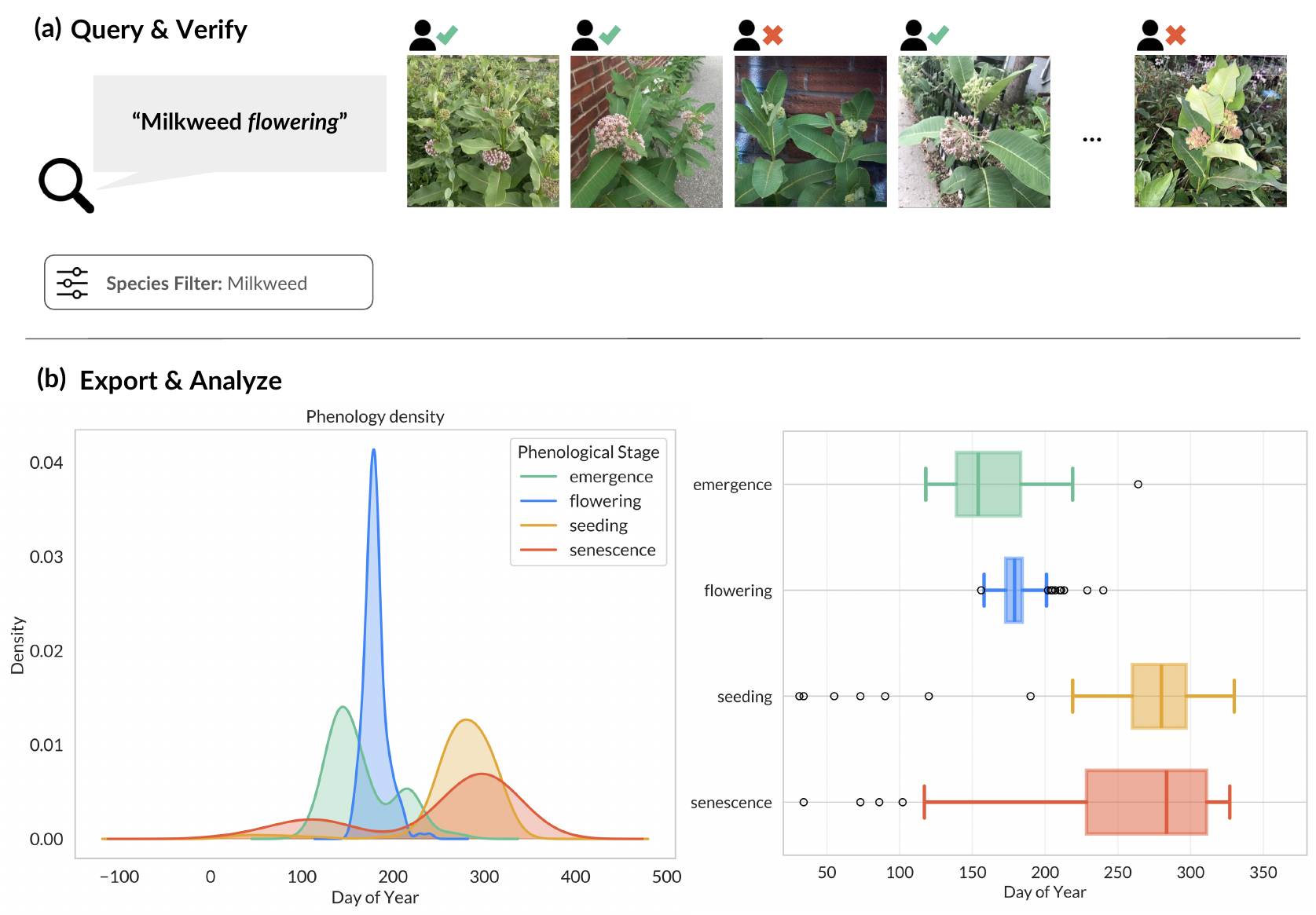}
    \vspace{-2mm}
    \caption{\textbf{(Top) Query \& Verification.} Filtering top INQUIRE-Search outputs for ``flowering'' phenological stage. \textbf{(Bottom) Phenological stages occur at significant times throughout the year.} (left) density of observations from INQUIRE-Search corresponding to different stages. (right) distribution of observations from INQUIRE-Search in different phenological stages visualized with results from ANOVA/Tukey test.}
    \label{fig:figure_10}
    \vspace{-2mm}
\end{figure}

\begin{figure}[h]
    \centering
    \includegraphics[width=\textwidth]{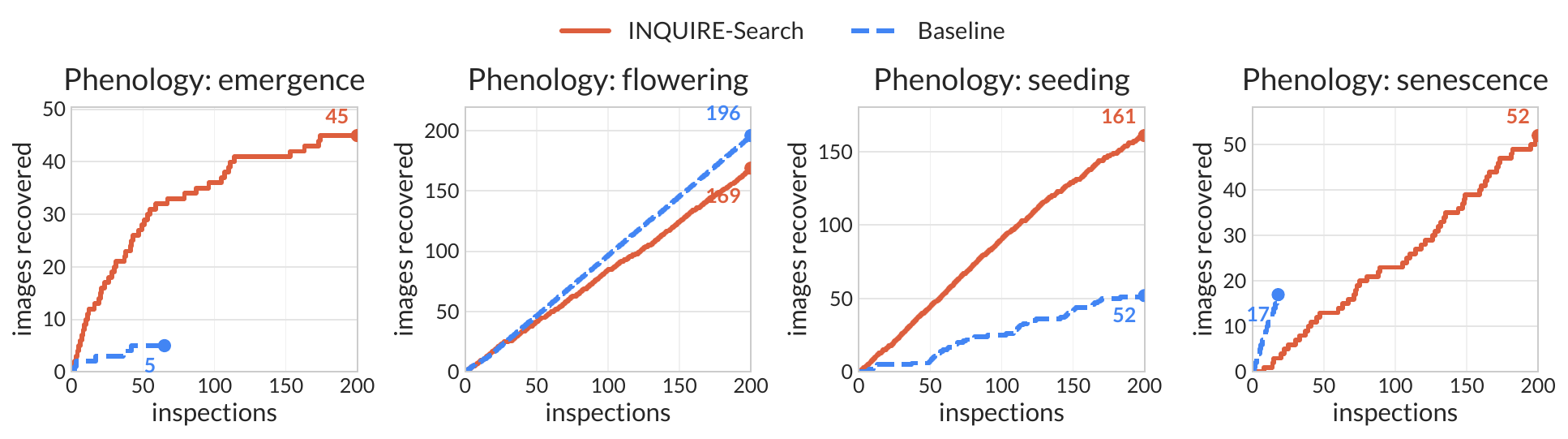}
    \caption{\textbf{Retrieval Curves for Plant Phenology.} Comparing recovered images vs inspections show INQUIRE-Search is generally more efficient and result in better yield, but when attributes are well-documented (e.g., flowering), metadata filtering is quite effective and is complementary to VLM-based search.}
    \label{fig:fig_11}
    \vspace{-4mm}
\end{figure}

In contrast, iNaturalist’s controlled phenology annotations performed unevenly across stages. We applied iNaturalist’s controlled phenology annotations, mapping “No Flowers or Fruits” to emergence, “Flowers” to flowering, “Fruits or Seeds” to seeding, and “No Live Leaves” to senescence. After verifying up to the first 200 images per stage, baseline filtering yielded 5 emergence ($Y=0.03$), 196 flowering ($Y=0.98$), 52 seeding ($Y=0.26$), and 17 senescence images ($Y=0.09$). \edit{The recovery curves in Figure 11 show that INQUIRE-Search reaches the metadata baseline quickly for emergence and seeding, while continuing to recover additional relevant observations for less consistently annotated stages. Flowering is an exception, where metadata annotations remain highly effective, suggesting that structured filters can complement VLM search when the target is visually salient and well-annotated.} Overall, structured metadata produced far fewer usable records for early and late phenophases ($Y_{ratio}=7.7$ and $2.9$ for emergence and senescence, respectively), whereas INQUIRE-Search recovered informative examples across all four stages. 

Using the verified INQUIRE-Search records, we observed a clear temporal progression across the growing season, with emergence occurring earliest, followed by flowering, seeding, and senescence (Figure 10, bottom), as expected. An ANOVA detected significant differences among stages in day-of-year distributions, and post-hoc Tukey tests indicated that all pairwise comparisons were significant (p < 0.05) except between emergence and flowering, which show substantial overlap. 

The datasets retrieved using INQUIRE-Search resolve fine-grained phenological structure at regional scales using community-science imagery. Emergence and senescence were more difficult to retrieve than flowering and seeding, reflecting subtle visual distinctions and observer biases toward photographing flowers and fruits \citep{iNaturalist202xHighQualityObservations}. Senescent observations distributed across much of the year, indicating delayed visibility that must be considered when interpreting phenological timing. Overall, these results demonstrate that INQUIRE-Search enables scalable, stage-specific phenological inference from opportunistic community-science imagery beyond what is feasible with traditional survey-based approaches.

\subsection{CS5: Whale re-identification}

\textbf{Overview.} {This case study evaluates whether INQUIRE-Search can surface individually identifiable humpback whale fluke images from unstructured community-science imagery and link them to known individuals in existing ID catalogues.} 

\begin{figure}[h]
    \centering
    \includegraphics[width=\textwidth]{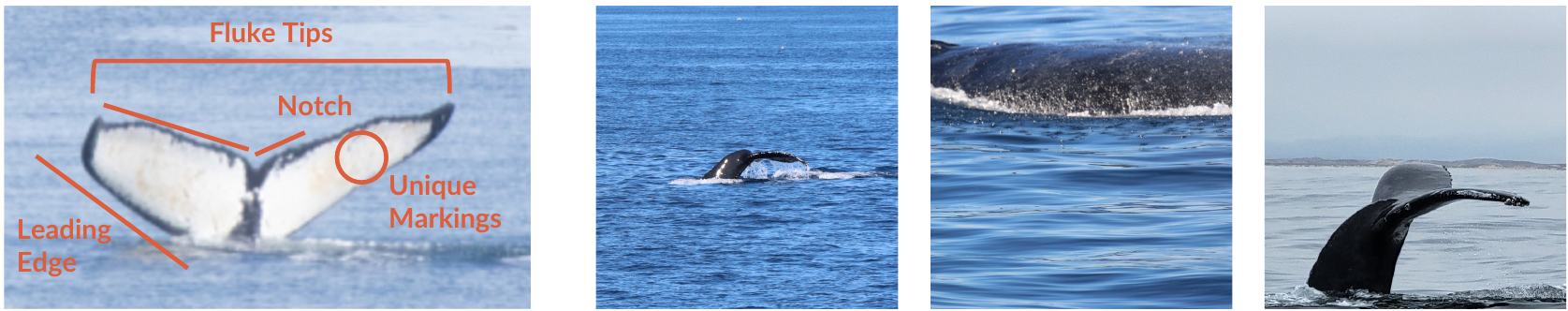}
    \caption{\textbf{(Left) Re-Identification of humpback whales.} Unique Re-Identification of individuals is possible through careful examination of the fluke. \textbf{(Right) iNaturalist humpback whale images with ``fluke'' description/tag filtering.} Recent iNaturalist humpback whale images, even with the fluke filtering, are not identifiable due to the photographed angle or image quality.}
    \label{fig:figure_12}
\end{figure}

Animal re-identification---recognizing individuals across space and time---is essential for population monitoring and movement ecology ~\citep{krebs1989ecological}. For wide-ranging marine mammals like humpback whales, these analyses rely on longitudinal photo-identification datasets that link repeated sightings using unique pigmentation and trailing-edge patterns on the ventral fluke, as shown in Figure 12 ~\citep{katona1981identifying, calambokidis2001movements, martin1984migration, howard2018humpback}. Although many fluke photographs exist in repositories such as iNaturalist, identifiable images are rare and dispersed across unstructured collections, limiting scalable re-ID dataset construction. We evaluate whether INQUIRE-Search can address this bottleneck by surfacing identifiable fluke images and linking them to known individuals in the HappyWhale dataset \citep{howard2018humpback}.

\textbf{{INQUIRE-Search workflow for CS5.}} \textbf{{1) Query and Prioritize:}} We queried for ``white underside of humpback whale fluke'' with the humpback whale species filter. \textbf{{2) Retrieve and Verify:}} Images were considered informative only if the ventral fluke was clearly visible, unobstructed, and in focus. Top $N_{insp}=200$ retrievals were inspected. \textbf{{3) Export and Analyze:}} For analysis, we assess whether any iNaturalist retrievals corresponded to known individuals in the HappyWhale dataset. To accelerate matching, we cropped each image using Grounding DINO \citep{liu2024grounding} and embedded both the iNaturalist and HappyWhale datasets using a multi-species re-identification model \citep{otarashvili2024multispecies}. For each of the verified INQUIRE-Search images, we retrieved the top three closest HappyWhale candidates by embedding similarity and manually reviewed these candidates to confirm matches. Although not conducted by experts, humpback flukes possess distinct and easily recognizable pigmentation and trailing-edge patterns that allow reliable identification by non-experts.

\textbf{{Results.}} INQUIRE-Search efficiently surfaced high-quality humpback whale fluke photographs suitable for individual identification (Figure 13). Of the $N_{insp}=200$ retrieved images examined, $N_{ret}=153$ images contained unobstructed, properly oriented, sufficient quality flukes, resulting in a screening yield of $Y=0.77$

For comparison, we used keyword filtering as a baseline, since no controlled term exists for fluke images on iNaturalist. We searched for “tail,” which yielded the largest candidate set, and verified the first $N_{insp}=200$ results, retaining $N_{ret}=42$ usable fluke images ($Y=0.21$). \edit{The recovery curve in Figure 14 shows that INQUIRE-Search ranks identifiable fluke images much earlier, matching the keyword baseline’s full yield after only 54 inspections.} Overall, INQUIRE-Search produced substantially more usable re-ID observations ($Y_{ratio}=3.67$), providing a stronger candidate set for downstream matching.

\begin{figure}[h]
    \centering
    \includegraphics[width=\textwidth]{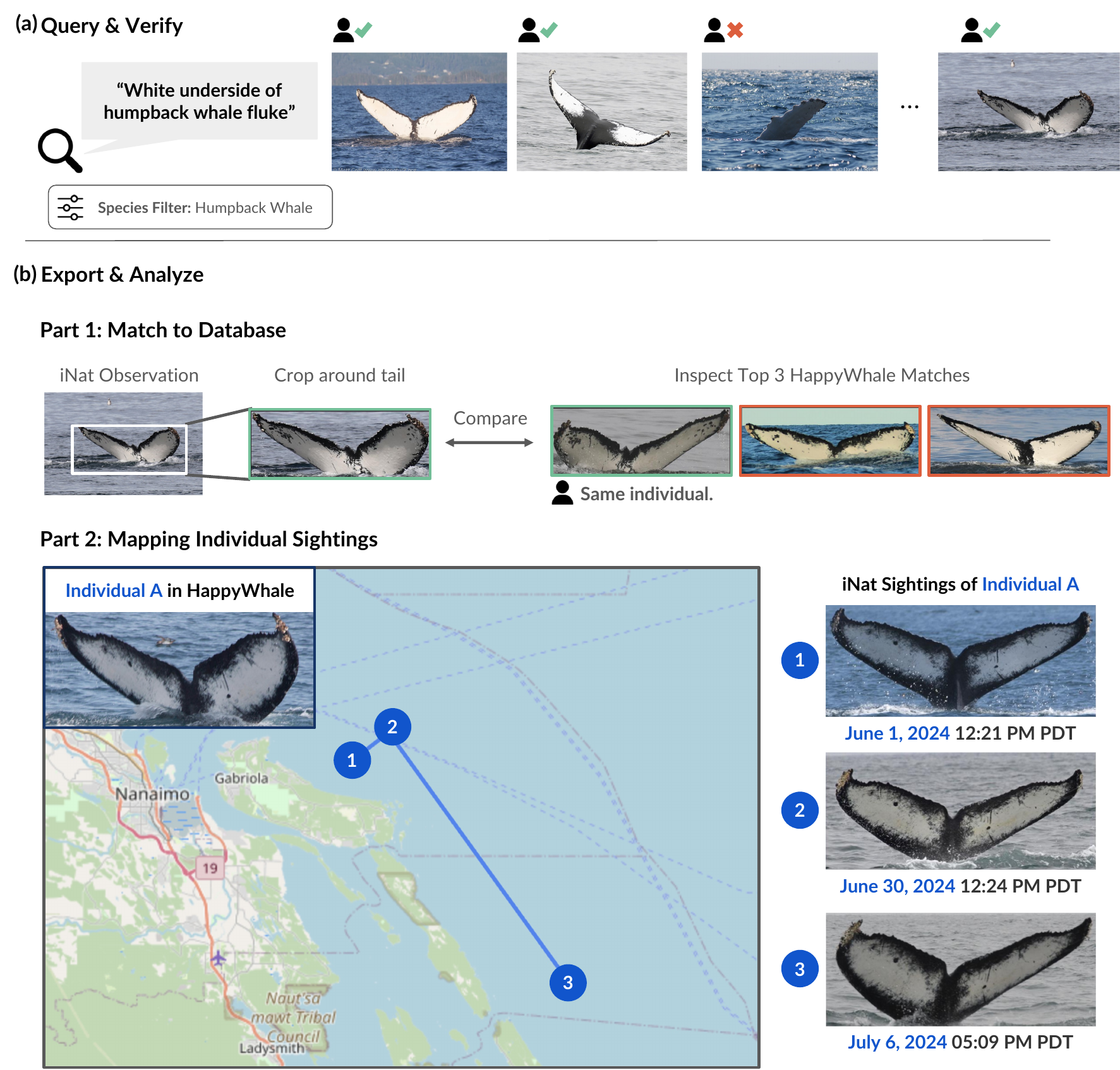}
    \vspace{-2mm}
    \caption{\textbf{(Top) Query \& Verification.} Filtering top INQUIRE-Search outputs for ``white underside of humpback whale fluke'' where the fluke was clearly visible and of sufficient quality for pattern matching. \textbf{(Bottom) Matching individuals to the HappyWhale database and analysis.} A deep learning Re-ID model retrieves candidate matches in the HappyWhale database, which we manually verify. We then map repeated sightings of confirmed individuals, demonstrating how INQUIRE-Search can support analyses of movement patterns and habitat use.}
    \label{fig:figure_13}
    \vspace{-2mm}
\end{figure}

\begin{figure}[h]
    \centering
    \includegraphics[width=0.5\textwidth]{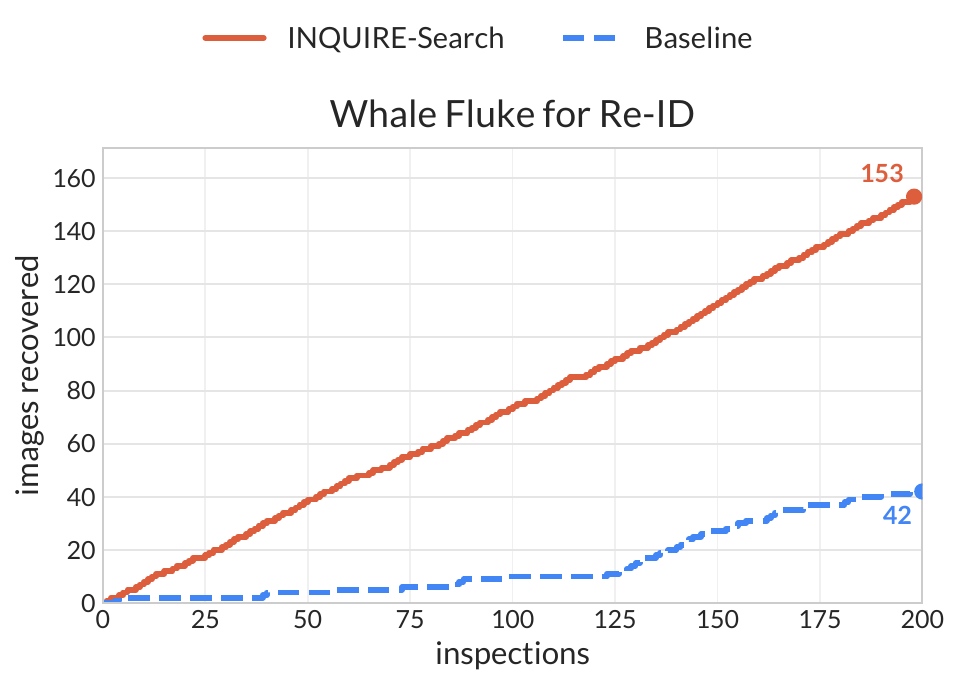}
    \caption{\textbf{Retrieval Curves for Whale Re-ID.} Comparing recovered images vs inspections show that INQUIRE-Search is both significantly more efficient and effective at prioritizing relevant images for whale fluke retrieval.}
    \label{fig:fig_14}
    \vspace{-4mm}
\end{figure}

Using the verified INQUIRE-Search images, we matched 57 observations to the HappyWhale dataset, corresponding to 34 unique humpback whales (Figure 13a). Several individuals appeared multiple times, extending their spatiotemporal coverage. Figure 13b visualizes matched iNaturalist and HappyWhale observations, highlighting how this linking expands our ability to track individuals across datasets. Since iNaturalist observations include geolocation metadata, cross-referencing matches provides new spatial information for individuals that were previously represented only in curated catalogs. This capability enables the identification of additional movement paths and unusual sightings.

INQUIRE-Search enables efficient discovery of correctly posed, identifiable fluke images from unstructured community-science imagery, facilitating scalable integration with existing photo-ID datasets. However, downstream identification remains dependent on pre-existing labeled catalogs and trained re-identification models. Also, retrieval reflects spatial and behavioral biases in community-sourced photography, favoring well-sampled locations and charismatic encounters over unbiased representations of whale distribution or behavior. 


\section{Discussion}
{INQUIRE-Search enables rapid, expert-guided discovery by efficiently surfacing analyzable ecological data from community-sourced images. 
The case studies presented in this work serve as concrete examples of how this new capability can be integrated into scientific workflows, illustrating how questions are posed, prompts refined, verification effort allocated, and downstream analyses conducted. Across diverse applications, we demonstrate that the system consistently recovers information that would otherwise require substantial manual effort or targeted fieldwork, helping to resolve temporal, spatial, and observational gaps.}

\vspace{-4mm}
\subsection{{INQUIRE-Search in practice: design considerations and limitations}}
{INQUIRE-Search reframes the early stages} of the scientific process by transforming large image archives from passive repositories into interactive substrates for reasoning. Scientists remain the drivers of inquiry, interpretation, and verification, but {the bottleneck shifts from data collection to validation.} Incorporating such a tool with an ecological research workflow requires rethinking experimental design. The following discussion highlights parameters to ``tune'' in this process, from identifying suitable questions and crafting precise prompts, to allocating verification effort and accounting for uncertainty.

\textbf{Which questions are well-posed for INQUIRE-Search?} INQUIRE-Search is fundamentally limited by the images that exist in the underlying repository. Questions targeting rare events, under-observed species, or obscure behaviors may yield insufficient data for robust analysis. For example, in the avian diet case study, species with few iNaturalist records produced too few relevant images of feeding behaviors to support quantitative comparisons across time. Conversely, when data are overly abundant, as in the wildlife mortality study, search still improves efficiency but does not eliminate the need for substantial verification. Useful queries must also be unambiguously visually verifiable: prompts tied to explicit, observable features (“milkweed with seed pods”) are more reliable than abstract states (“tree under stress”). Well-posed questions therefore target phenomena that are (1) sufficiently represented in the data pool, (2) not so broad that they produce large volumes results, and (3) visually verifiable.

\textbf{How to refine a question: iterating on prompts.} Search-based discovery depends on clear and specific prompts. Precise language reduces ambiguity in the embedding space and improves the relevance of returned images. Specificity also increases the verifiability of the outputs: ``robin pulling a worm from the soil'' will yield cleaner results than ``bird foraging.'' Iteration is often necessary, where prompts can be refined by adding behavioral, contextual, or anatomical cues, while overly narrow prompts can be relaxed when returns are sparse. \edit{We further discuss some potential effects of prompt variability in the Appendix \ref{sec:appendix_rephrasing}.}

\textbf{Managing effort, ranking, and verification.} This workflow introduces a new design choice: in addition to formulating effective queries, researchers must also decide how far down the ranked search results to inspect. In this way, ``effort'' becomes tunable. Reviewing only the top-ranked subset often recovers the most relevant observations with minimal redundancy, as shown in the dietary analysis case study, where experts verified only the first 200 images recovered to obtain a sufficient quantity of data.



\textbf{Identifying uncertainty and bias with INQUIRE-Search.} INQUIRE-Search inherits the biases of community-science platforms. Observer bias skews data toward charismatic, accessible, or unusual events, while cryptic interactions and remote habitats remain under-represented \citep{dimson2023observer}. \edit{To address this, while querying data, researchers can use temporal, spatial, and taxonomic stratification to reduce obvious imbalances. During analysis, retrieved observations can be paired with effort proxies and adjusted using observer-effort correction, weighting, or adaptive sampling approaches \citep{robinson2018correcting, guilbaultexplorers, mondain2024adaptive, padilla2024assessing}.}

\edit{Model-driven biases add a second layer of uncertainty. VLMs can struggle with relational or compositional prompts \citep{alhamoud2025visionlanguagemodelsunderstandnegation, thrush2022winogroundprobingvisionlanguage}, be miscalibrated \citep{guo2017calibration}, or rely on spurious correlations rather than visual evidence \citep{vo2025vision}. These uncertainties can be integrated into the INQUIRE-Search workflow: selective prediction can prioritize low-confidence or underrepresented cases for expert review \citep{geifman2017selective}; and uncertainty quantification can help propagate error analysis into downstream scientific reports \citep{narduzzi2014inverse, karimi2023quantifying}.}

{INQUIRE-Search supports dataset aggregation rather than end-to-end statistical inference. With experts in the loop, search outputs are treated as hypotheses rather than ground truth, placing responsibility on researchers to communicate uncertainty and limitations downstream.}

{\subsection{Computation cost}}
{INQUIRE-Search is built on embedding and retrieval rather than repeated model inference, substantially reducing computational and energy costs once the system is deployed. Images in the databases are embedded by the AI model once using a pretrained VLM and stored in an index, and each search requires only a \textit{single} embedding of the user's text query (single model inference call) followed by fast similarity matching. In contrast, other AI models such as generative models often reprocess \textit{all images and text in the database} for each query, requiring repeated, compute-intensive inference that scales poorly with dataset size. The marginal energy cost of interactive discovery with INQUIRE-Search is low, making it a practical approach to large-scale search and retrieval.}

\subsection{\edit{Future Extensions}}
Although our case studies drew on iNaturalist for its scale and ecological breadth, the open-source design of the INQUIRE-Search codebase makes it adaptable to any large image collection. {The system’s novelty lies in the mechanism of discovery rather than the data source,} enabling scientists, NGOs, and local communities to deploy the same open-source and resource-efficient workflow to harness image search for ecological monitoring and hypothesis generation on their own databases. \edit{INQUIRE-Search is also model-agnostic, and its performance should improve as ecological retrieval models advance, with approaches such as biodiversity-trained VLMs \citep{stevens2024bioclip, gabeff2024wildclip}, hierarchical ecological concept embeddings \cite{stevens2024bioclip}, and ontology-guided query expansion \citep{gomes2007ecologically, nguyen2017constructing, amanqui2013semantic}.}


\section{Conclusion}

Growing ecological image databases remain an underutilized scientific resource. INQUIRE-Search, our proposed workflow for efficient, open-ended data discovery and verification via text-based search, enables scalable analysis of diverse ecological phenomena, from bird diet to plant phenology. By helping scientists extract ecological evidence from community-sourced imagery, it maximizes the value of existing datasets, provides a low-resource starting point for hypothesis testing, and helps prioritize resource-intensive data collection campaigns to fill remaining gaps. However, any new experimental methodology requires best practices to ensure rigorous, reproducible, and unbiased analyses. Our case studies show the diversity of potential applications and demonstrate the efficiency with which experts can \edit{convert hypotheses into curated datasets that can support exploratory analysis and motivate formal inference}: this offers a scalable pathway to accelerate ecological insight in a rapidly changing world. {By reducing discovery time across vast image databases, the system enables faster hypothesis generation and prioritization of follow-up studies. While search-based discovery does not replace systematic surveys, \edit{it can reveal patterns, gaps, and candidate signals that help prioritize targeted data collection and formal hypothesis testing.}}

\subsubsection*{Acknowledgments} 
We appreciate all the global contributors to the iNaturalist platform
for their collection of species observations that make up the backbone
of this work. We also want to thank the iNaturalist team and the authors
of the INQUIRE benchmark, particularly Scott Laurie, Alex Shepard,
Grant Van Horn, Omiros Pantazis, Gabriel Brostow, and Kate Jones.
This work was supported in part by a Schmidt Science’s AI2050 Early
Career Fellowship, NSF CAREER Grant (Award No. 2441060), an NSF
Graduate Research Fellowship (Award No. DGE-2146755), the NSF and
NSERC AI and Biodiversity Change Global Center (NSF Award No.
2330423 and NSERC Award No. 585136), and the MIT Generative AI
Consortium. Oisin Mac Aodha was in part supported by a Royal Society
Research Grant. LLMs were used minimally in the preparation of this
manuscript, limited to light editing and clarity checks, with all scientific
content authored by the researchers.

\subsubsection*{Glossary}
\textbf{Cosine Similarity}: A metric measuring angular similarity between vectors, commonly used to rank semantic relevance.\\\
\textbf{Embedding Vector}: A numerical representation of an image or text query used for similarity search and other downstream tasks.\\
\textbf{Model Bias}: Systematic errors introduced by training data or model assumptions that affect retrieval results.\\
\textbf{Natural-Language Query}: A free-text query written in plain language rather than predefined labels or categories.\\
\textbf{Open-Set Retrieval}: A retrieval setting where queries are not restricted to predefined classes or labels.\\
\textbf{Opportunistic Sampling}: Data collection driven by observer behavior rather than structured survey design.\\
\textbf{Phenology}: The timing of recurring biological events (e.g., flowering, senescence) across seasonal cycles.\\
\textbf{Sampling Bias}: Non-random representation of locations, species, or behaviors in collected data.\\
\textbf{Secondary Data}: Ecologically relevant information captured in images beyond species presence, such as behavior, interactions, or habitat context.\\
\textbf{Vector Similarity Search}: A retrieval method that ranks items based on distance (e.g., cosine similarity) between embedding vectors.\\
\textbf{Vision–Language Model (VLM)}: A machine learning model that embeds images and text into a shared semantic space, enabling cross-modal search and comparison.

\subsubsection*{Declaration of competing interest}
The authors declare that they have no known competing financial
interests or personal relationships that could have appeared to influence the work reported in this paper.

\subsubsection*{Declaration of generative AI and AI-assisted technologies in the manuscript preparation process}
During the preparation of this work the author(s) used GPT-4o in order to catch grammatical errors and other inconsistencies in the writing. After using this tool/service, the authors reviewed and edited the content as needed and take full responsibility for the content of the published article.

\subsection*{Appendix A: INQUIRE-Search system architecture and design motivation}
\label{sec:appendixa}
INQUIRE-Search is a tool designed to enable scientists to quickly and easily discover data from within a large ecological database such as iNaturalist. The system architecture of INQUIRE-Search reflects both of these priorities, combining state-of-the-art vision-language models with efficient indexing and memory management techniques to deliver responsive searches across hundreds of millions of images on modest hardware (4 vCPUs with 32GB RAM, roughly the capacity of a single laptop).

\subsubsection*{Search index}

 First we embed text query using the same SigLIP model used to embed the iNaturalist images. We then calculate the vector similarity between that new text embedding and each image embedding and use the score to rank the full set of images to surface images of interest. With a smaller database, this entire process can be accomplished directly within the memory of a modest computer, but when scaling to the hundreds of millions of images in iNaturalist the set of image embeddings becomes large ($\sim$200GB), requiring a different approach for a rapid new search.

Vector databases, which use a specialized indexing system that allows fast search over millions of high-dimensional vectors, \citep{decastro2018detecting, singla2021experimental, fang2017topology} are specifically designed to facilitate efficient similarity-based search. We utilize FAISS \citep{douze2025faiss} to create an approximate nearest neighbor index that is small and fast to query. Our FAISS index is generated and tuned using the Autofaiss library, a widely-used library for efficient nearest-neighbor search  \citep{webster2023duplication}. The final index is memory-mapped (stored on disk but accessed as if it were in memory, enabling fast search with limited RAM), uses 36GB of storage, and enables sub-500ms search times while maintaining high retrieval accuracy.

In addition to the visual embedding index, our search tool supports several metadata-based mechanisms to filter by:
\begin{enumerate}
    \item \textbf{Taxonomy:} We map each taxon (species, genus, family, etc.) to a list of image IDs known to contain that entity, using taxon labels provided for each image from iNaturalist. This allows for efficient subset selection when users apply taxonomic filters to their queries.
    \item \textbf{Temporal Range:} Users can filter search queries by month.
    \item \textbf{Geographic Extent:} Using approximate image location data, optionally provided by iNaturalist contributors, users can filter images to a specific geographic area using latitude-longitude bounds. 
\end{enumerate}

\subsubsection*{Image and text embeddings}
The core search capability is powered by a vision-language model. Not all vision-language models are equally capable of ranking ecological data, particularly for scientific queries. We leveraged our previous work developing the INQUIRE-Benchmark, which specifically evaluates models for ecological image retrieval tasks, to select SigLIP-So400m-384-14 \citep{zhai2023sigmoid} as our embedding model, due to both its strong performance on the benchmark and its reasonable tradeoff between speed and accuracy; other models like ViT-H-14 \citep{radford2021learning} are slightly more accurate at retrieving relevant data but are significantly larger, and thus slower. We processed 300M images sourced from iNaturalist through the visual encoder component of the selected SigLIP model to build the backbone of the INQUIRE-Search tool. 

\begin{figure}[h]
    \centering
    \includegraphics[width=\textwidth]{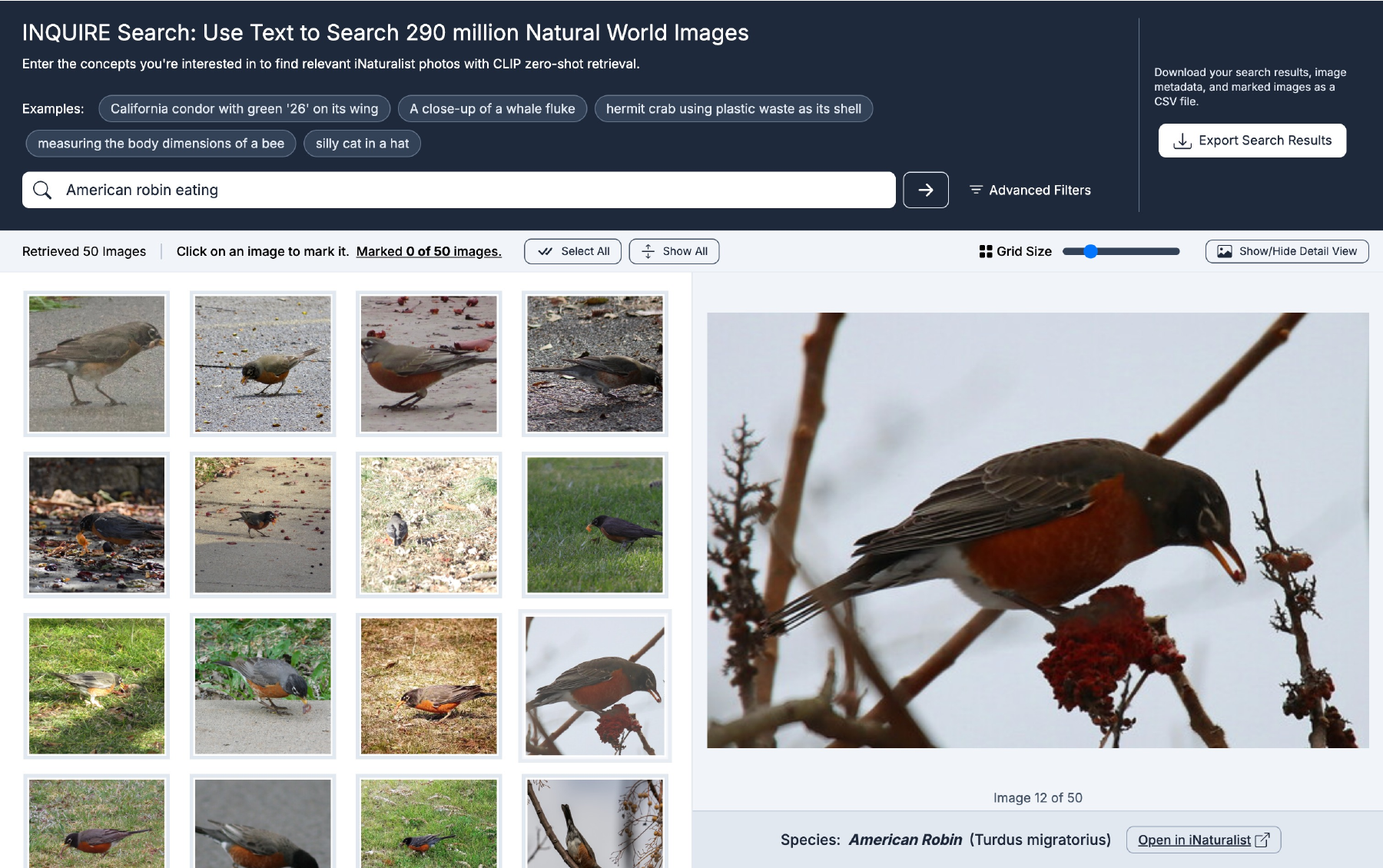}
    \caption{INQUIRE-Search provides an user-friendly and intuitive interface that allows ecologists to query hundreds of millions of natural-world images. This example illustrates a query for ``American robin eating,'' with species and temporal filters applied, enabling rapid retrieval, inspection, and verification of relevant observations.}
    \label{fig:figure_15}
\end{figure}

\subsubsection*{Search interface}
INQUIRE-Search provides a web-based interface (Figure 15) that surfaces the search index and filters to scientists. The interface includes a query field for natural language queries (e.g., ``California condor with a green ‘26’ tag on its wing'') and controls for taxonomic, temporal, or geospatial filters. The interface displays a grid of thumbnails of the images returned from the user-provided query, ordered by relevance to the query. The images are clickable, allowing the user to mark images for further analysis. An expanded view of a selected image shows a full-resolution image with complete metadata.

\subsubsection*{Data export}

When the user exports the data, a CSV file is prepared for download. This file contains a row for each image which includes a field for whether an image was marked by the user, detailed metadata, and a link to the associated iNaturalist observation.

\subsubsection*{Data analysis}

Once verified, these retrieved observations form the foundation for a wide range of statistical analyses, demonstrated in the following section. This workflow allows experts to effectively ``design'' their search strategy and specify what ecological signals to target, while identifying or mitigating relevant sources of uncertainty and bias.

\subsection*{Appendix B: Retrieval and filtering results}
\label{sec:appendixb}

\begin{table}[htbp]
    \centering
    \scriptsize
    \setlength{\tabcolsep}{3pt}
    \begin{tabularx}{\textwidth}{l X l l X l}
        \toprule
        \textbf{Species} & \textbf{Prompt Template} & \textbf{Season} & \textbf{Inspected} & \textbf{Filtered} & \textbf{Rate} \\
         & & & & \tiny \textit{(Inv/Vert/Seed/Fruit/Nect/Carr/Plant)} & \\
        \midrule
        \multirow{2}{*}{\textbf{American Robin}}
         & \multirow{10}{=}{``\textless species\textgreater\ with \textless diet type\textgreater\ in its mouth''}
         & Summer & 500 per type & 338 & 67.6\% \\
         & & Winter & 500 per type & 256 & 51.2\% \\
        \cmidrule{1-1} \cmidrule{3-6}
        \multirow{2}{*}{\textbf{Red-bellied Woodpecker}}
         & & Summer & 500 per type & 95 & 19.0\% \\
         & & Winter & 500 per type & 161 & 32.2\% \\
        \cmidrule{1-1} \cmidrule{3-6}
        \multirow{2}{*}{\textbf{American Tree Sparrow}}
         & & Summer & 500 per type & 1 & 0.2\% \\
         & & Winter & 500 per type & 83 & 16.6\% \\
        \cmidrule{1-1} \cmidrule{3-6}
        \multirow{2}{*}{\textbf{Ancient Murrelet}}
         & & Summer & 500 per type & 0 & 0.0\% \\
         & & Winter & 500 per type & 2 & 0.4\% \\
        \cmidrule{1-1} \cmidrule{3-6}
        \multirow{2}{*}{\textbf{Gray-cheeked Thrush}}
         & & Summer & 500 per type & 0 & 0.0\% \\
         & & Winter & 500 per type & 0 & 0.0\% \\
        \multicolumn{6}{p{\textwidth}}{\tiny \textbf{Key:} Inv=Invertebrate, Vert=Vertebrate, Nect=Nectar, Carr=Carrion. Counts are placeholders (-/...) where detailed breakdowns were not provided in source text. Total inspected per diet/season/species is 500.}
    \end{tabularx}
    \vspace{2mm}
    \caption{Retrieval results for seasonal variation in bird diets. The prompt template is applied to all species. Diet types are abbreviated in the header row.}
    \label{tab:cs1_birds}
\end{table}

\begin{table}[htbp]
    \centering
    \small
    \begin{tabularx}{\textwidth}{l X X c c c}
        \toprule
        \textbf{Target} & \textbf{Prompt} & \textbf{Filters} & \textbf{Insp.} & \textbf{Filt.} & \textbf{Rate} \\
        \midrule
        Young coniferous trees 
        & ``young coniferous trees in burned forest'' 
        & \multirow{2}{=}{Geo: High Park Fire (40.57-40.75$^{\circ}$ N, 105.18-105.54$^{\circ}$ W); Date: Post-2012}
        & 100 & 45 & 45.0\% \\
        \cmidrule{1-2} \cmidrule{4-6}
        Young deciduous trees 
        & ``young deciduous trees in burned forest'' 
        & 
        & 112 & 78 & 69.6\% \\
        \bottomrule
        \multicolumn{6}{l}{\footnotesize Filters sourced from Table 2.}
    \end{tabularx}
    \caption{Search parameters and results for post-fire forest regrowth.}
    \label{tab:cs2_forest}
\end{table}

\begin{table}[htbp]
    \centering
    \small
    \begin{tabularx}{\textwidth}{l X X c c c}
        \toprule
        \textbf{Location} & \textbf{Prompt} & \textbf{Filters} & \textbf{Insp.} & \textbf{Filt.} & \textbf{Rate} \\
        \midrule
        Boston (Urban) & ``dead bird'' & Geo: Bounding box around Boston, MA; No Taxon filter & 1000 & 543 & 54.3\% \\
        \midrule
        Pioneer Valley (Rural) & ``dead bird'' & Geo: Bounding box around Pioneer Valley/Amherst, MA; No Taxon filter & 360 & 79 & 21.9\% \\
        \bottomrule
    \end{tabularx}
    \caption{Search parameters and results for wildlife mortality.}
    \label{tab:cs3_mortality}
\end{table}

\begin{table}[htbp]
    \centering
    \small
    \begin{tabularx}{\textwidth}{l X X c c c}
        \toprule
        \textbf{Stage} & \textbf{Prompt} & \textbf{Filters} & \textbf{Insp.} & \textbf{Filt.} & \textbf{Rate} \\
        \midrule
        Emergence & ``Milkweed germinating or emerging'' & \multirow{4}{=}{Taxon: \textit{Asclepias syriaca}; Geo: S. Quebec (45.03 to 46.54$^{\circ}$ $^{\circ}$ N, -74.68 to -71.66$^{\circ}$ W); No temporal filter} & 200 & 45 & 22.5\% \\
        Flowering & ``Milkweed flowering'' & & 200 & 169 & 84.5\% \\
        Seeding & ``Milkweed producing seeds...'' & & 200 & 161 & 80.5\% \\
        Senescence & ``Milkweed dying or withering...'' & & 200 & 52 & 26.0\% \\
        \bottomrule
    \end{tabularx}
    \caption{Search parameters and results for plant phenology stages.}
    \label{tab:cs4_phenology}
\end{table}

\begin{table}[htbp]
    \centering
    \small
    \begin{tabularx}{\textwidth}{l X X c c c}
        \toprule
        \textbf{Target} & \textbf{Prompt} & \textbf{Filters} & \textbf{Insp.} & \textbf{Filt.} & \textbf{Rate} \\
        \midrule
        Humpback whale flukes & ``white underside of humpback whale fluke'' & Taxon: Humpback Whale (\textit{Megaptera novaeangliae}); No Geo/Temp filters & 200 & 153 & 76.5\% \\
        Humpback whale flukes & ``white underside of humpback whale fluke'' & Taxon: Humpback Whale (\textit{Megaptera novaeangliae}); No Geo/Temp filters & 200 & 153 & 76.5\% \\
        \bottomrule
    \end{tabularx}
    \caption{Search parameters and results for humpback whale re-identification.}
    \label{tab:cs5_whale}
\end{table}

\newpage
\subsubsection*{\edit{Effect of Prompt Rephrasing on Retrieval Performance}}
\label{sec:appendix_rephrasing}
\edit{We evaluated prompt rephrasing for CS1: American Robin summer diet retrieval and found that wording affected retrieval yield. For fruit, the common-name prompt, ``American robin holding fruit in its beak,'' performed best, yielding 201 images per 500 inspected, while similar scientific-name prompts (``Turdus Migratorius with fruit in its mouth,'', ``Turdus Migratorius feeding on fruit,'') yielded 148 and 163 images. For invertebrates, the same common-name phrasing also performed best, over scientific-name prompts. These differences suggest that users should test multiple phrasings and, when aiming for more exhaustive labeling, stop only after a sustained run of non-relevant images rather than after a fixed inspection budget. Prompt choice should therefore be treated as a practical source of retrieval bias and an experimental design choice.}
\newpage 
\clearpage

\bibliographystyle{apalike}
\bibliography{main}

\end{document}